\documentclass{article}
\usepackage{ijcai24}

\usepackage{times}
\usepackage{soul}
\usepackage{url}
\usepackage[hidelinks]{hyperref}
\usepackage[utf8]{inputenc}
\usepackage[small]{caption}
\usepackage{graphicx}
\usepackage{amsmath}
\usepackage{amsthm}
\usepackage{booktabs}
\usepackage{algorithm}
\usepackage{algorithmic}
\usepackage[switch]{lineno}

\usepackage{paralist}
\usepackage{color}
\usepackage{xcolor}
\usepackage{subcaption}
\usepackage{multirow}
\usepackage{acronym}
\usepackage{bm}
\usepackage{listings,lstautogobble}
\newcommand{\asp}[1]{\mbox{$\mathtt{#1}$}}

\newcommand{\method}{NeSyGPT}

\acrodef{method}[NeSyGPT]{NeSyGPT}
\acrodef{asp}[ASP]{Answer Set Programming}
\acrodef{nesy}[NeSy]{Neuro-Symbolic AI}
\acrodef{llm}[LLM]{Large Language Model}
\acrodef{las}[LAS]{Learning from Answer Sets}
\acrodef{vqa}[VQA]{Visual Question Answering}

\DeclareMathOperator*{\argmin}{\arg\min}

\title{The Role of Foundation Models in Neuro-Symbolic Learning and Reasoning}

\author{
Daniel Cunnington$^{1,2}$\and
Mark Law$^3$\and
Jorge Lobo$^4$\And
Alessandra Russo$^2$\\
\affiliations
$^1$IBM Research Europe\\
$^2$Imperial College London\\
$^3$ILASP Limited\\
$^4$ICREA-Universitat Pompeu Fabra\\
\emails
dtc20@imperial.ac.uk, mark@ilasp.com, jorge.lobo@upf.edu, a.russo@imperial.ac.uk
}

\begin{document}

\maketitle

\begin{abstract}
    \ac{nesy} holds promise to ensure the safe deployment of AI systems, as interpretable symbolic techniques provide formal behaviour guarantees. The challenge is how to effectively integrate neural and symbolic computation, to enable learning and reasoning from raw data. Existing pipelines that train the neural and symbolic components sequentially require extensive labelling, whereas end-to-end approaches are limited in terms of scalability, due to the combinatorial explosion in the symbol grounding problem. In this paper, we leverage the implicit knowledge within foundation models to enhance the performance in \ac{nesy} tasks, whilst reducing the amount of data labelling and manual engineering. We introduce a new architecture, called \method{}, which fine-tunes a vision-language foundation model to extract symbolic features from raw data, before learning a highly expressive answer set program to solve a downstream task. Our comprehensive evaluation demonstrates that \method{} has superior accuracy over various baselines, and can scale to complex \ac{nesy} tasks. Finally, we highlight the effective use of a large language model to generate the programmatic interface between the neural and symbolic components, significantly reducing the amount of manual engineering required.
\end{abstract}

\section{Introduction}

Ensuring the safe deployment of AI systems is a top-priority. \acf{nesy} \cite{GarcezGLSST19} is well-suited to help achieve this goal, as interpretable symbolic components enable manual inspection, whilst providing formal guarantees of their behaviour. However, integrating neural components to enable learning and reasoning from raw data remains an open challenge. Existing approaches either require extensive data labelling and manual engineering of symbolic rules \cite{eiter2022neuro}, or, when neural and symbolic components are trained end-to-end \cite{ijcai2021-254}, they encounter difficulties scaling to complex tasks. This is due to the combinatorial explosion in learning the correct assignment of symbolic features to the raw input (i.e., the symbol grounding problem \cite{harnad1990symbol}). End-to-end approaches are also difficult to control, causing unintended concepts to be learned, thus posing a challenge to AI safety \cite{nesyconcepts}. Pre-trained foundation models on the other hand, have a large amount of implicit knowledge embedded within their latent space, which could be leveraged during the training of a \ac{nesy} system. The question we address in this paper is: \textit{How can we use this knowledge to reduce the amount of data labelling and engineering required, whilst solving complex neuro-symbolic tasks that have many possible values for the symbolic features?} 

We introduce a new architecture, called \method{}, that fine-tunes the BLIP vision-language foundation model \cite{li2022blip} to extract symbolic features from raw data, before learning a set of logical rules to solve a downstream task. The architecture utilises the implicit knowledge of BLIP, and therefore, we only require a few labelled data points to fine-tune for a specific task. Logical rules are learned as \ac{asp} programs, suited for efficiently representing solutions to computationally complex problems \cite{gelfond2014knowledge}. The rules are learned using a state-of-the-art \ac{las} symbolic learner robust to noise. We evaluate \method{} in four problem domains, each with specific challenges, and demonstrate that it out-performs a variety of neural and \ac{nesy} baselines. The main contribution of this paper is a novel architecture that integrates a vision-language foundation model and a symbolic leaner for solving complex neuro-symbolic tasks, offering the following advantages:

\begin{enumerate}
\item It requires a reduced amount of labelled data points to extract symbolic features from raw data and learn the rules of the downstream task.
\item It scales to complex tasks that have a large number of possible symbolic feature values.
\item It does not require the symbolic rules to be manually engineered.
\item It can solve tasks that require detecting multiple objects and their properties within a single image, e.g., the CLEVR-Hans3 dataset \cite{stammer2021right}. 
\item It can utilise a \ac{llm} to generate questions and answers for fine-tuning BLIP, and to construct the programmatic interface between the neural and symbolic components.
\end{enumerate}

\begin{figure*}[t]
    \centering
    \begin{tabular}[t]{cc}
    \begin{subfigure}[c]{0.36\textwidth}
        \centering  
        \includegraphics[keepaspectratio, width=\textwidth]{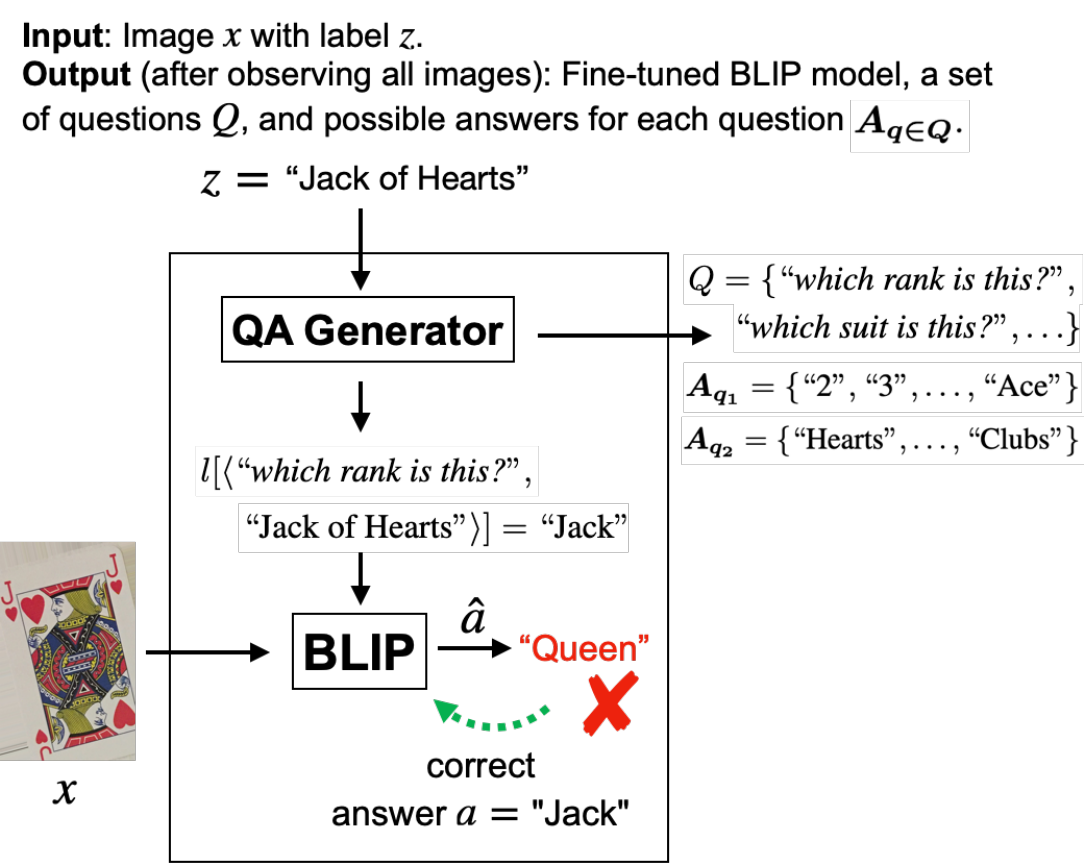}
        \caption{BLIP Fine-Tuning}
        \label{fig:blip_ft}
    \end{subfigure}
    \hfill
    \hspace{2em}
    \hfill
    \begin{tabular}{c}
        \begin{subfigure}[t]{0.57\textwidth}
            \centering
            \includegraphics[keepaspectratio, width=\textwidth]{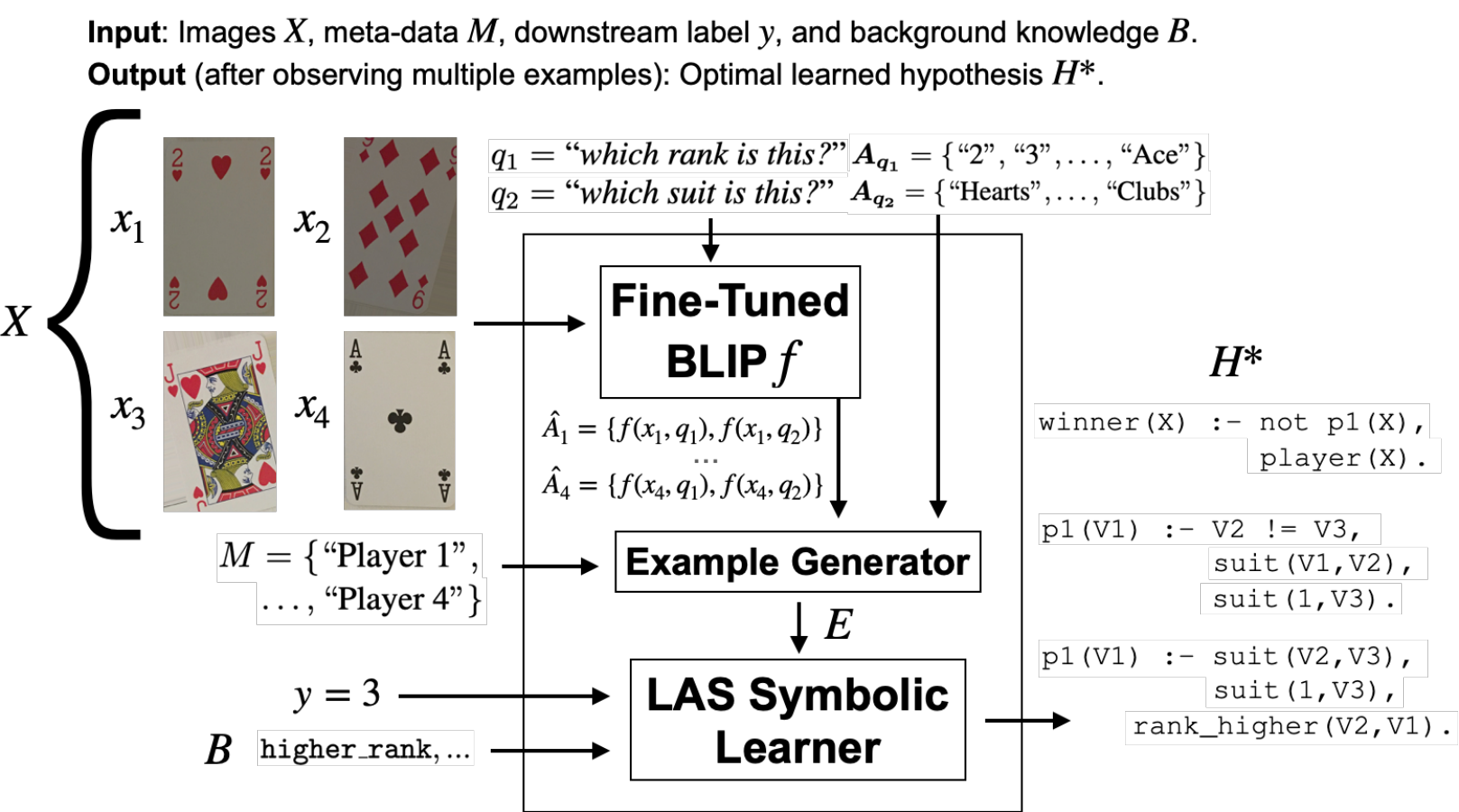}
            \caption{Symbolic Rule Learning}\label{fig:learning}
        \end{subfigure}\\
    \end{tabular}\\
    \end{tabular}
    \caption{\method{} architecture with one data point in the Follow Suit task. The goal is to learn the rules of the game: The winner is the player with the highest ranked card with the same suit as Player 1. (a) BLIP is fine-tuned using playing card images and natural language questions and answers. (b) A hypothesis is learned from BLIP predictions. Note in (a), fine-tuning occurs with both suit and rank answers.}
    \label{fig:arch}
\end{figure*}

\section{Related Work}
There are various \ac{nesy} systems proposed in the literature \cite{Besold2017}. Some inject neural architectures with symbolic rules \cite{riegel2020logical,Badreddine2022}, whereas others perform neural and symbolic computation in separate modules \cite{manhaeve2018deepproblog,neurasp,aspis2022embed2sym,cunnington2023neuro,ijcai2023p461}. Whilst various architectures, types of logic, and computation techniques have been explored, many \ac{nesy} systems lack the expressivity required to learn or reason with general and complex rules, and are either restricted to definite clauses \cite{manhaeve2018deepproblog,dai2019bridging,ijcai2021-254,shindo2023alpha}, graphical models with cost matricies \cite{ijcai2023p402}, or look-up tables \cite{ijcai2023p400}. In this paper, we focus on \ac{asp}-based \ac{nesy} systems, as they are able to express general and complex rules. 

Current \ac{nesy} approaches still suffer drawbacks. 
Some rely on pre-trained pipelines that require extensive data labelling and manual engineering \cite{eiter2022neuro,cunnington2023ffnsl}, whereas others are limited in terms of scalability, in the case of end-to-end training \cite{neurasp,EVANS2021103521,aspis2022embed2sym,cunnington2023neuro,skryagin2023scalable,charalambous2023neuralfastlas}. This is due to the aforementioned symbol grounding problem. Furthermore, only \cite{EVANS2021103521,cunnington2023ffnsl,cunnington2023neuro,charalambous2023neuralfastlas} are able to perform rule learning from raw data in \ac{asp} for \ac{nesy} tasks, whereas other approaches assume the symbolic rules are manually specified. Our approach is able to seamlessly integrate rule learning and reasoning in \ac{asp}, and scales to tasks with a large amount of choice for the values of the symbolic features, thus avoiding the combinatorial explosion that end-to-end methods suffer. Also, we require very few labelled data points, and a reduced amount of manual engineering, thus gaining an advantage over existing pipeline approaches. Finally, it is difficult to integrate object detection within current \ac{nesy} architectures, as they train a CNN to classify an entire image. Our approach is more general, and can extract various objects and their properties from a single image through \ac{vqa} (see CLEVR-Hans results in Section \ref{sec:results}).

Integrating foundation models with symbolic computation has also been studied previously. \cite{yang2023coupling,KR2023-37} leverage \acp{llm} to generate \ac{asp} programs to solve complex reasoning problems, and \cite{nye2021improving} use a \ac{llm} to iteratively generate candidate solutions, accepting a solution if it satisfies some given background knowledge. In contrast to our approach, these focus purely on language models, and are unable to support visual input. Also, none of these approaches learn symbolic rules, as they are either specified symbolically, or in textual form.

There are cases where rules are generated automatically. \cite{wang2023hypothesis} uses a language model to perform inductive reasoning, by generating multiple candidate hypotheses and having a human-oracle or another language model select a subset for validation. This technique does not support visual input, and crucially, the final hypothesis is chosen as the candidate that covers the most training examples, during a post-processing stage. In contrast, our approach takes advantage of a symbolic learner that includes a more advanced optimisation algorithm, which maximises example coverage as part of the hypothesis search, and provides a formal guarantee of the optimality and consistency of the learned rules. Finally, \cite{suris2023vipergpt} achieves \ac{vqa}, by using a \ac{llm} to generate python programs that call external modules for image processing. However, as evidenced in our results, it is difficult to generate solutions to complex problems that require negation, choice, and constraints using current \acp{llm} \cite{kassner2019negated}. Our approach can learn and efficiently express such rules using \ac{asp}.

\section{Background}\label{sec:bk}
\textbf{BLIP}. BLIP uses a multi-modal mixture of encoder-decoder transformer modules. The image encoder is a Vision Transformer, and the text encoder is a BERT model. BLIP is pre-trained to jointly optimise three objectives; (1) \textit{Image-Text Contrastive Loss}, which aligns the embedding space of the vision transformer and BERT by encouraging positive image-text pairs to have similar latent representations in contrast to negative pairs. (2) \textit{Image-Text Matching Loss}, which is a binary classification task that trains BLIP to predict whether an image-text pair is matched (positive) or unmatched (negative). (3) \textit{Language Modelling Loss} which aims to generate a textual description given an image. This is achieved using the cross-entropy loss with a ground-truth description. In this paper, we use the ``\ac{vqa}" mode to perform fine-tuning. The pre-trained BLIP model is rearranged by encoding an image-question pair into multi-modal embeddings, before passing the embeddings to a text decoder to obtain the answer. The \ac{vqa} architecture is then trained end-to-end using the Language Modelling Loss with ground-truth answers. For more details, please refer to \cite{li2022blip}.

\textbf{\ac{las}}. A \ac{las} symbolic learner learns a set of rules, called a \textit{hypothesis}. The symbolic learner accepts as input a set of training examples $E=\{\langle e_{\text{id}}, e_{\text{pos}}, e_{\text{pen}}, e_{\text{inc}}, e_{\text{exc}}, e_{\text{ctx}} \rangle,\ldots\}$, where $e_{\text{id}}$ is an identifier, $e_{\text{pos}}$ denotes whether the example is either positive or negative, $e_{\text{pen}}$ is a weight penalty, which is paid if the example is not \textit{covered} by the learned rules, $e_{\text{inc}}$ and $e_{\text{exc}}$ contain inclusion and exclusion atoms respectively, and $e_{\text{ctx}}$ contains a set of contextual facts. Alongside $E$, the symbolic learner also requires a domain knowledge $B$ that contains relations used to construct a search space $\mathcal{H}$, and any (optional) background knowledge. The output is a hypothesis $H\in \mathcal{H}$ in the form of an \ac{asp} program. $H$ minimises a scoring function based on the length of $H$, and its example coverage. Let us assume the set of examples uncovered by $H$ is denoted by $UNCOV(H,(B,E))$. The score of $H$ is given by $score(H,(B,E)) = \vert H \vert +  \sum_{e\in UNCOV(H,(B,E))}{e_{\text{pen}}}$. A \ac{las} symbolic learner computes the optimal solution $H^{*}$ that satisfies Equation \ref{lasequation}:

\begin{equation}
\label{lasequation}
H^{*} = \argmin_{H\in \mathcal{H}}\; [ score(H, (B,E)) ]
\end{equation}
This can be interpreted as jointly maximising the generality of $H$ (i.e., a concise \ac{asp} program), and example coverage. For more details, please refer to \cite{LawRB19}.
\section{Method}
















\subsection{Problem Formulation}

We assume a perception function $f:\langle\mathcal{X},\mathcal{Q}\rangle \rightarrow \mathcal{A}$ that performs \ac{vqa}. $\mathcal{X}$, $\mathcal{Q}$, and $\mathcal{A}$ are the spaces of raw image inputs, natural language questions, and answers, respectively. The answers represent symbolic features related to an image. We also assume a reasoning function $h:\langle\mathcal{A}^{u},\mathcal{M}\rangle^{n} \rightarrow \mathcal{Y}$ that maps $n$-length sequences of answers and meta-data pairs to a target value. Note that there could be $u$ questions asked of each image, and therefore, $u$ answers for each pair. The meta-data in the space $\mathcal{M}$ is used to encode any extra information associated with the raw input. $\mathcal{Y}$ is the space of the target values. 

During training, we use two datasets: (1) $D_{\text{image}}=\{\langle x,z\rangle,\ldots\}$, where $x\in\mathcal{X}$ is an image, and $z\in \mathcal{Z}$ is an image label, and (2) $D_{\text{task}}=\{\langle X,M,y\rangle,\ldots\}$ where $X$ is a sequence of images $x\in\mathcal{X}$, $M\subseteq \mathcal{M}$ is a (possibly empty) set of meta-data associated with $X$, and $y\in\mathcal{Y}$ is a downstream task label. We learn $f$ and $h$ sequentially,\footnote{where $h$ is learned using predictions from $f$, once $f$ is learned.} as our experiments demonstrate this outperforms an end-to-end approach (see Section \ref{sec:results}). To learn $f$, we require, in addition to the raw input, sets of questions and answers. These are generated automatically, and are assumed constant for a given task. We denote the set of questions as $Q \subseteq \mathcal{Q}$ and the set of possible answers for a given question $q\in Q$ as $\boldsymbol{A_{q}}\subseteq \mathcal{A}$. 
For example, in the Follow Suit task (see Figure \ref{fig:arch}), the question $q=$``\textit{which rank is this?}" has the set of possible answers $\boldsymbol{A_{q}}=\{``2",``3",\ldots,``\text{Ace}"\}$. We also require a look-up table $l:\langle \mathcal{Q}, \mathcal{Z} \rangle \rightarrow \mathcal{A}$ which is used to map question and image label pairs to the correct answer. For example, given an image label $z=\text{``Jack of Hearts"}$, the look-up table returns the answer ``Jack" for the question $q=\text{``which rank is this?"}$, and the answer ``Hearts" for $q=\text{``which suit is this?"}$.

The goal is to learn $f$ s.t. $\forall \langle x,z\rangle \in D_{\text{image}}$, and $\forall q \in Q$, $f(x,q) = l(\langle q, z \rangle)$, i.e., $f$ returns the correct answer for all images and all questions. Once $f$ is learned, the next objective is to learn $h$. Let us define the set of predicted answers for an image $x_{i} \in X$ as $\hat{A_{i}} = [f(x_{i}, q_{1}),\ldots,f(x_{i}, q_{u})]$, where $q_{j}\in Q$. The goal is to learn $h$, s.t. $\forall \langle X,M,y\rangle \in D_{\text{task}}$, $h([\langle \hat{A_{i}}, m_{i} \rangle: x_{i}\in X]) = y$, where $m_{i}\in M$. Intuitively, this means that each data point in $D_{\text{task}}$ has a correct downstream prediction. For example, in Figure \ref{fig:arch}, assuming correct predictions, the answers are $\hat{A_{1}}= [\text{``2"}, \text{``Hearts"}]$, $\hat{A_{2}}= [\text{``9"}, \text{``Diamonds"}]$, $\hat{A_{3}}= [\text{``Jack"}, \text{``Hearts"}]$, and $\hat{A_{4}}= [\text{``Ace"}, \text{``Clubs"}]$, and the meta-data $M$ indicates which player each card corresponds to, e.g., $m_{1} = ``\text{Player 1}"$. The target label is $y=3$, as Player 3 is the winner. We propose the \method{} architecture which consists of the BLIP Fine-Tuning phase for learning $f$, and the Symbolic Rule Learning phase for learning $h$ (see Figure \ref{fig:arch}).

\subsection{BLIP Fine-Tuning}
Given an image data point $\langle x_{i}, z_{i} \rangle \in D_{\text{image}}$, and a question $q \in Q$, the answer returned from the look-up table $l(\langle q, z_{i} \rangle)$ is used to fine-tune BLIP, alongside the image $x_{i}$. 
The question-answer generator is specific to each task, and can be manually engineered or obtained automatically using a \ac{llm}. Our experiments show that the \ac{llm} output is below the level of accuracy achieved with manual engineering. See the Supp. Material for a detailed analysis. However, we expect this to improve as \acp{llm} themselves continue to improve. To fine-tune BLIP with the generated questions and answers, we use the ``\ac{vqa}" mode as presented in Section \ref{sec:bk}. 


\subsection{Symbolic Rule Learning}
\textbf{Example Generator}. Given a task data point $\langle X,M,y\rangle \in D_{\text{task}}$, and a sequence of answer meta-data pairs $[\langle \hat{A_{i}}, m_{i} \rangle: x_{i}\in X]$ from the BLIP Fine-Tuning phase, the goal is to generate a LAS training example $e \in E$ for the symbolic learner. The inclusion and exclusion sets of $e$ encode the downstream label. Formally, $e_{\text{inc}}=\{ y \}$, and $e_{\text{exc}}=\{ y^{\prime} : y^{\prime} \in \mathcal{Y}, y^{\prime} \neq y \}$. The answers and meta-data are encoded into the example's context $e_{\text{ctx}}$. In the case of feature classification questions (e.g., $q=\text{``which rank is this?"}$), there is no guarantee that BLIP predicts a valid answer for a given question. To tackle this problem, we use the Levenshtein string distance to return the answer $a \in \boldsymbol{A_{q}}$ with the lowest edit distance compared to the predicted answer $\hat{a}$ \cite{yujian2007normalized}. The answers and meta-data are then encoded into $e_{\text{ctx}}$ in \ac{asp} form, which is task specific. For example, in Figure \ref{fig:learning}, assuming correct BLIP predictions, $e_{\text{ctx}} = \{\asp{card(1,2,h).}, \asp{card(2,9,d).}, \asp{card(3,j,h).}, \asp{card(4,a,c).} \}$, representing the ``\textit{Two of Hearts}", ``\textit{Nine of Diamonds}", ``\textit{Jack of Hearts}", and ``\textit{Ace of Clubs}" for each player's card, respectively. The Example Generator can be manually engineered, or obtained using a \ac{llm}, which generates Python code that encodes the \ac{asp} representation of a full training example $e$. We use the string distance method to obtain valid answers before using the LLM to encode $e$. In the Supp. Material, we demonstrate that with small syntactic fixes, the \ac{llm} can encode the training examples with competitive accuracy compared to a manual specification.

\textbf{\ac{las} Symbolic Learner}
Given a set of training examples $E$, and a domain knowledge $B$, the symbolic learner learns an optimal hypothesis $H^{*}$, satisfying the scoring function in Equation \ref{lasequation}. $H^{*}$, together with $B$, forms the reasoning function $h=H^{*} \cup B$. Let us now present our experimental results.

\section{Evaluation}\label{sec:results}
Our evaluation investigates whether the implicit knowledge embedded within BLIP can improve the accuracy of \ac{nesy} learning and reasoning, whilst reducing the amount of labelled data required.\footnote{Code will be released upon paper acceptance.} We compare \method{} against the following types of architectures: \textit{\ac{nesy} Reasoning}, where the symbolic rules are given, \textit{\ac{nesy} Learning}, where the symbolic rules are learned, and \textit{Fully Neural Methods} including deep learning, and generative models.

We choose four problem domains; MNIST Arithmetic, Follow Suit, Plant Disease Hitting Sets, and CLEVR-Hans, each with their own characteristics. The MNIST Arithmetic domain includes the standard \ac{nesy} benchmark from \cite{manhaeve2018deepproblog}, and the E9P task from \cite{cunnington2023neuro}, where we demonstrate learning negation as failure. The Follow Suit domain tests scalability w.r.t. the number of choices for the symbolic features \cite{cunnington2023ffnsl}. A given data point has $\sim6.5$m choices in the 4-player variant, and $\sim5.74\times 10^{16}$ choices in the 10-player variant. For reference, the 2-digit MNIST Addition task has 100 choices per data point. This domain also requires learning predicates that are not directly observed in the examples (i.e., predicate invention). The Plant Hitting Sets domain includes real-world images, and requires learning complex rules including constraints and choice, as well as reasoning over multiple models (answer sets). Finally, the CLEVR-Hans domain tests object detection, as up to 10 objects and their properties need to be extracted from a single image.

The results show that \method{}: (1) Achieves better performance than any of the other approaches. (2) Requires very few labelled data points. (3) Can scale to complex tasks with a large number of choices for the symbolic features. (4) Can easily integrate symbolic learning of \ac{asp} programs to learn complex and interpretable rules. (5) Can achieve state-of-the-art performance on CLEVR-Hans. We also demonstrate that the amount of manual engineering can be reduced using GPT-4 \cite{openai2023gpt4}. The results indicate GPT-4 can generate reasonable questions and answers for feature extraction. 
Due to space limitations, we present these results in the Supp. Material.

\textbf{Baselines.} Our baselines have symbolic components that are either fully neural, or are capable of learning or representing complex rules expressed in \ac{asp}. This is in order to show that the proposed benefits of \method{} are due to the implicit knowledge embedded within BLIP, and our novel method of integrating BLIP with the symbolic component, as opposed to the symbolic representation used. 

For \ac{nesy} Reasoning, we compare our approach to \textbf{NeurASP} \cite{neurasp}, \textbf{SLASH} \cite{skryagin2023scalable}, and two variants of Embed2Sym \cite{aspis2022embed2sym}: One trained end-to-end as usual (\textbf{Embed2Sym}), and another where clustering is performed on an embedding space resulting from fine-tuning a perception model with ground-truth image labels (\textbf{Embed2Sym*}). We also use a modified version of ABL \cite{dai2019bridging} as an additional baseline, which uses the same perception component as in our approach (\textbf{BLIP+ABL}). Note that BLIP can't directly be used in NeurASP, SLASH, and Embed2Sym because these methods require back-propagating gradients to a softmax layer for training the perception model. BLIP on the other hand, is fine-tuned with natural language questions and answers. 
For NeurASP, SLASH, and Embed2Sym, we use the best pre-trained model available; the Vision Transformer \textit{ViT-B-32} from PyTorch, which is pre-trained on ImageNet.

For \ac{nesy} Learning, we compare to \textbf{FF-NSL }\cite{cunnington2023ffnsl} and \textbf{NSIL} \cite{cunnington2023neuro}, using the pre-trained ViT-B-32 as the perception model. For Fully Neural Methods, we compare to: (1) The ViT-B-32 connected to a fully differentiable reasoning network (\textbf{ViT+ReasoningNet}), which is an MLP in the MNIST Arithmetic and Follow Suit domains, and a Recurrent Transformer in the Plant Hitting Sets domain, since longer sequences are observed \cite{hutchins2022block}. (2) BLIP connected to GPT-4 for rule learning (\textbf{BLIP+GPT4}). We prompt GPT-4 to generate Python programs for the MNIST Arithmetic and Follow Suit domains, since it is easy to express solutions to these tasks in Python, and GPT-4 has likely seen many Python examples during pre-training. For Plant Hitting Sets, we prompt GPT-4 to generate \ac{asp} programs, given it is more natural to express the solution to this problem in a declarative manner. (3) BLIP solving the downstream task with all images contained within a single image (\textbf{BLIP Single Image}). Finally, in the CLEVR-Hans domain, we consider \textbf{\boldsymbol{$\alpha$}ILP} \cite{shindo2023alpha} as our baseline, since $\alpha$ILP has achieved state-of-the-art performance.

\textbf{Experiment Setup.} In the MNIST Arithmetic, Follow Suit, and Plant Hitting Sets tasks, we generate questions and answers for fine-tuning BLIP using a varying number of image labels: $\{0,1,2,5,10,20,50,100\}\times$ the number of feature values for each image. To learn the symbolic rules, we use $\{10,20,50,100\}$ labelled examples for the symbolic learner. In all datasets, we use an even distribution of labels. For the baselines, we use the same total number of labels, distributed as follows. For NeurASP, SLASH, Embed2Sym, BLIP+ABL, NSIL, and BLIP Single Image, we train these models end-to-end with the same total number of labels as required for our approach, with no supervision on the symbolic features. For Embed2Sym*, FF-NSL, ViT+ReasoningNet, and BLIP+GPT4, we follow a similar approach to ours: We first fine-tune the perception component using ground-truth image labels, and then train the reasoning component using the same number of task labels. 


In the MNIST Arithmetic and CLEVR-Hans domains, we use FastLAS for the symbolic learner due to FastLAS’ increased efficiency in larger search spaces for these specific class of problems \cite{law2020fastlas}. We use ILASP in the Follow Suit and Plant Hitting Sets domains, as ILASP supports predicate invention (required in the Follow Suit domain), and the ability to learn choice rules and constraints (required in the Plant Hitting Sets domain) \cite{law2020ilasp}. We set the \ac{las} example weight penalty to 10 in all experiments except CLEVR-Hans, where a penalty of 1 is used. 10 and 1 were chosen because they provide adequate balance against the length of the hypothesis. In CLEVR-Hans, there are more symbolic examples given, so a lower weight penalty is required to learn a shorter, more general hypothesis.

Each experiment is repeated 5 times using 5 randomly generated seeds and datasets. We use the ``vqav2" BLIP model from the LAVIS library \cite{li2022lavis}, which is pre-trained on the Visual Genome, Conceptual Captions, Conceptual 12M, SBU captions, LAION, and the VQA v2.0 datasets. In our paper, BLIP is fine-tuned using a distinct set of training images, and the images in the training examples for the symbolic learner are not used for BLIP fine-tuning. We use 5 held-out test sets (not seen in any training stage) to evaluate task accuracy, and report the mean and standard error. For all models, we train with the most performant number of epochs we can achieve within a 24 hour time limit. For \method{}, we train for 10 epochs on MNIST Arithmetic and Follow Suit, and 20 epochs on Plant Hitting Sets and CLEVR. For all baselines, we train for 50 epochs unless otherwise stated. All hyper-parameters are left as default, and we use the SGD optimiser with a learning rate of 0.01 for all models, unless specified in a methods codebase. All experiments were run on a shared cluster with the following specification: RHEL 8.5 x86 64 with Intel Xeon E5-2667 CPUs (10 cores
total), and 1 NVIDIA A100 GPU, 150GB RAM. The domain knowledge and search spaces are given in the Supp. Material.

\subsection{MNIST Arithmetic}
This includes 2-digit Addition from \cite{manhaeve2018deepproblog}, and E9P, where the input is 2 MNIST images, and the label is equal to the digit value of the second image, if the digit value of the first image
is even, or 9 plus the digit value in the second image otherwise. The search space for the symbolic learner follows \cite{cunnington2023neuro} and includes the relations $\asp{even}$, $\asp{not \ even}$, $\asp{plus\_nine}$, and `=', as well as the function `+'. We fine-tune BLIP using the question ``\textit{which digit is this?}", and use the corresponding digit value as the answer. Image accuracies are reported over the standard MNIST test set of 10,000 images, and task accuracies are reported over held-out test sets of 1000 examples.

\begin{figure*}[t]
    \centering
    \begin{subfigure}[t]{0.8\textwidth}
            \centering
            \includegraphics[width=1\linewidth]{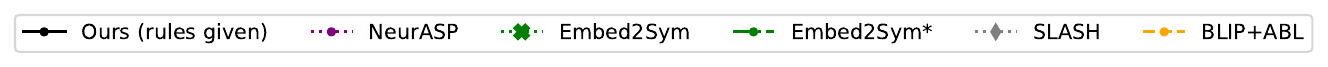}
        \end{subfigure}
        
        \begin{subfigure}[t]{0.24\textwidth}
            \includegraphics[width=\textwidth]{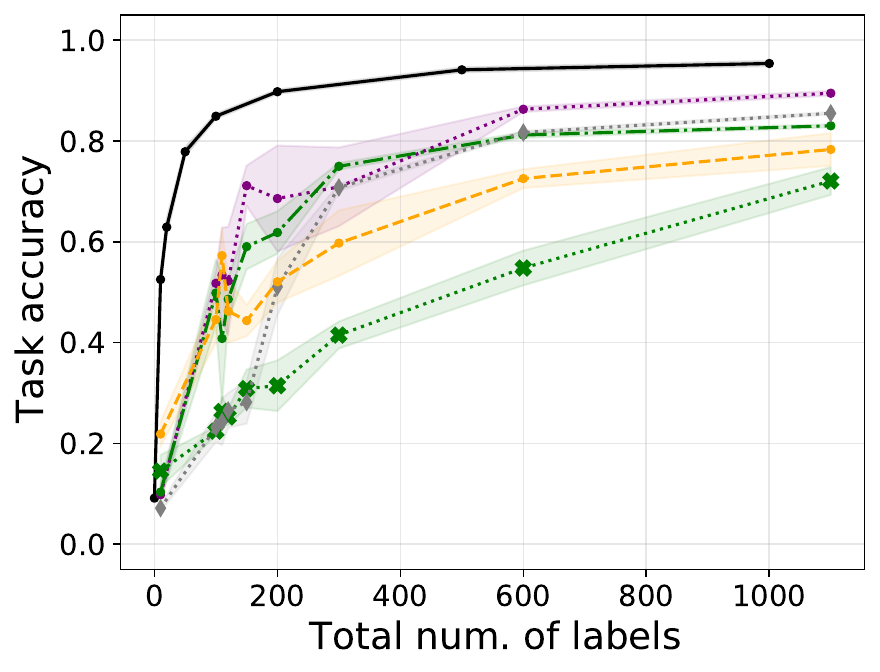}
            \caption{Addition}
            \label{fig:addition_reasoning}
        \end{subfigure}
        \begin{subfigure}[t]{0.24\textwidth}
            \includegraphics[width=\textwidth]{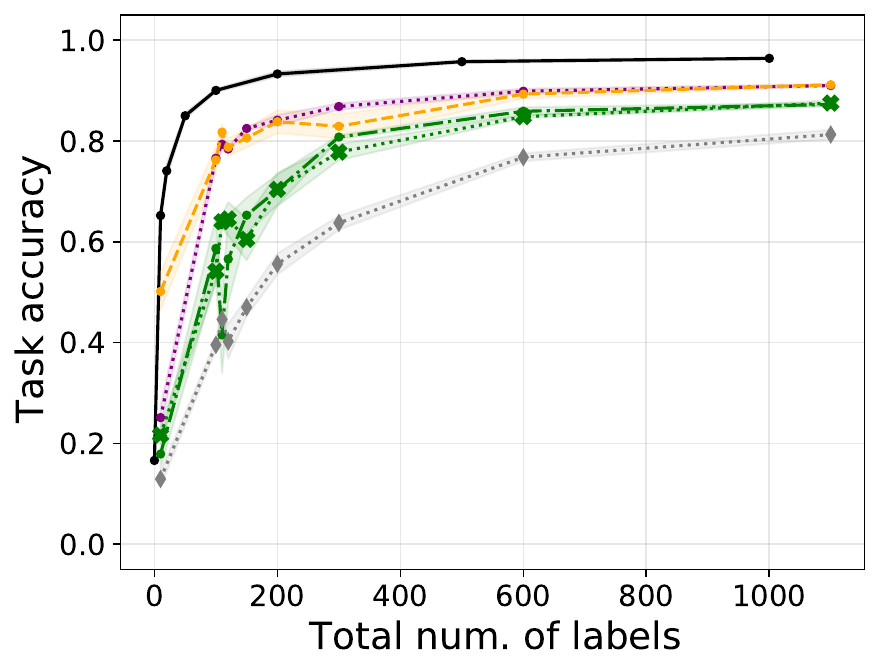}
            \caption{E9P}
            \label{fig:e9p_reasoning}
        \end{subfigure}
        \begin{subfigure}[t]{0.24\textwidth}
            \includegraphics[width=\textwidth]{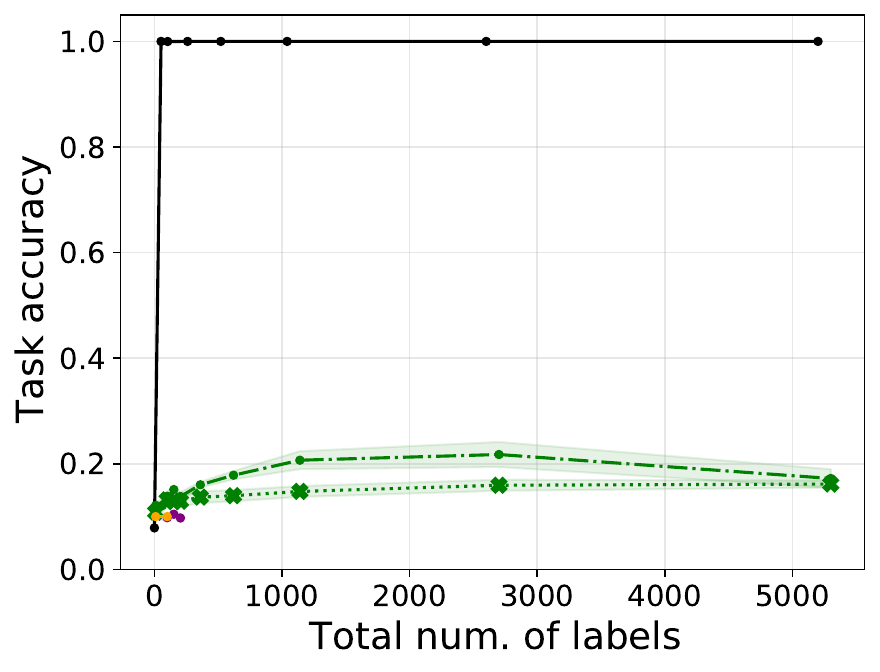}
            \caption{Follow Suit 10 ACA}
            \label{fig:fs_10_aca_reasoning}
        \end{subfigure}
        \begin{subfigure}[t]{0.24\textwidth}
            \includegraphics[width=\textwidth]{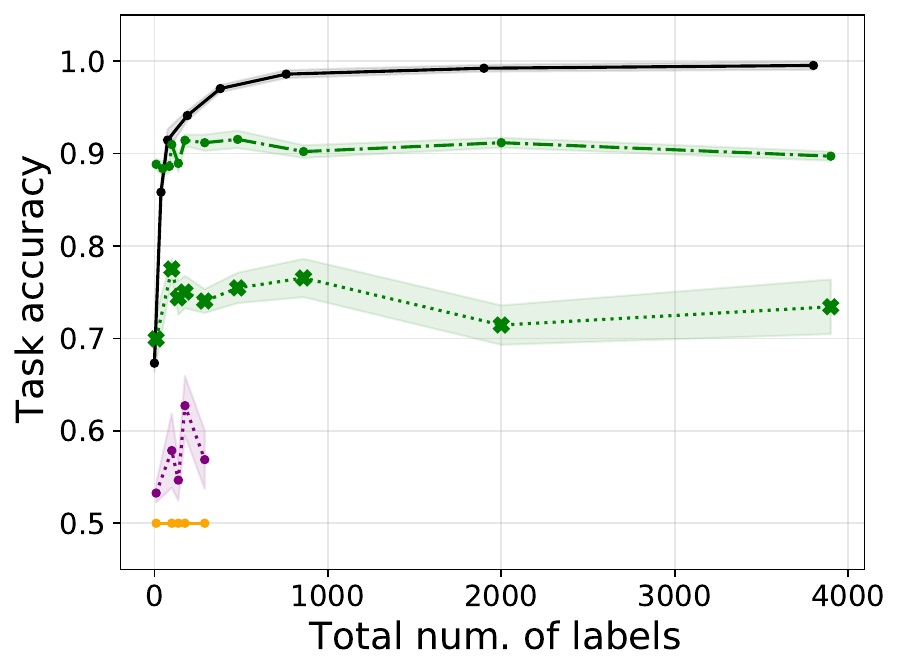}
            \caption{Plant Hitting Sets}
            \label{fig:plant_hs_reasoning}
        \end{subfigure}
        \caption{Task accuracy when the symbolic rules are given.}
        \label{fig:reasoning_results}
\end{figure*}

\begin{figure*}[t]
    \centering
    \begin{subfigure}[t]{0.8\textwidth}
            \centering
            \includegraphics[width=1\linewidth]{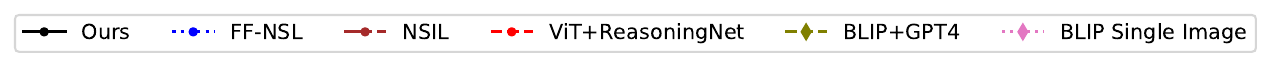}
        \end{subfigure}
        
        \begin{subfigure}[t]{0.24\textwidth}
            \includegraphics[width=\textwidth]{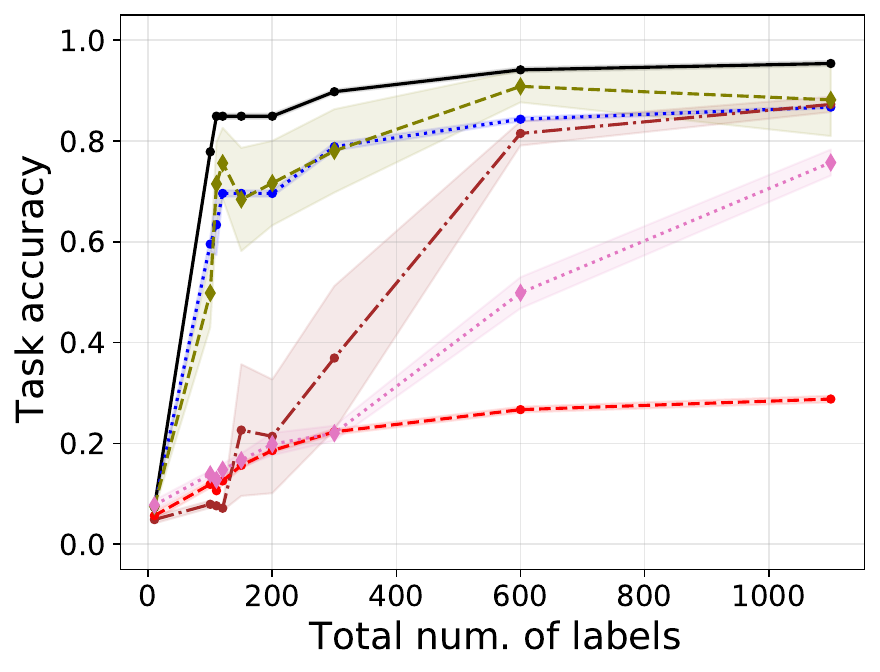}
            \caption{Addition}
            \label{fig:addition_learning}
        \end{subfigure}
        \begin{subfigure}[t]{0.24\textwidth}
            \includegraphics[width=\textwidth]{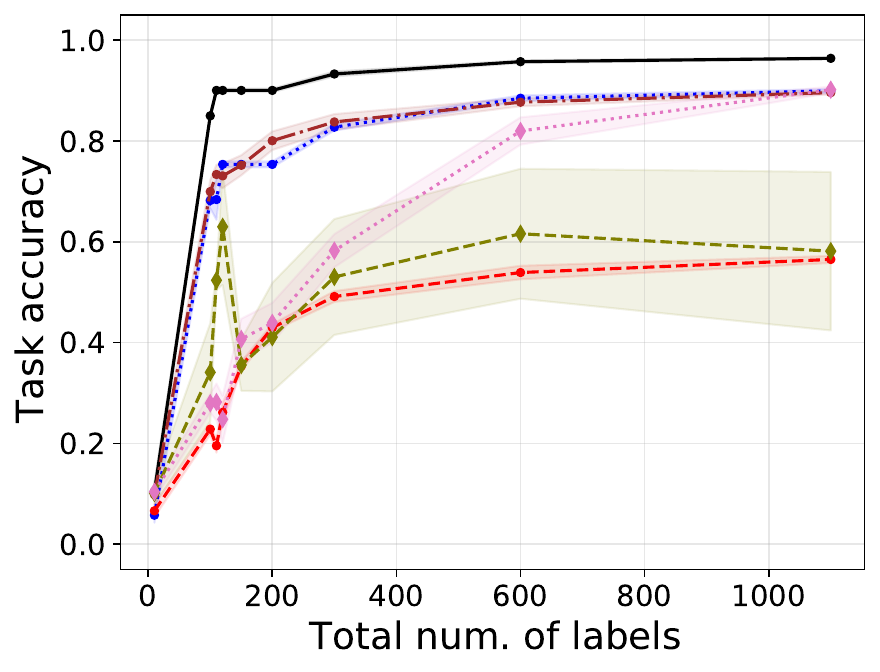}
            \caption{E9P}
            \label{fig:e9p_learning}
        \end{subfigure}
        \begin{subfigure}[t]{0.24\textwidth}
            \includegraphics[width=\textwidth]{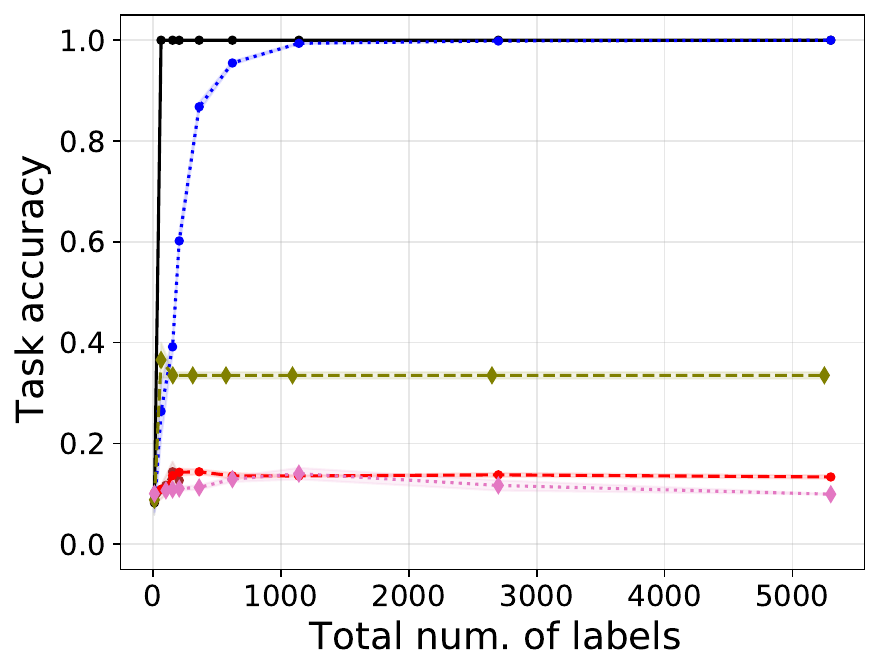}
            \caption{Follow Suit 10 ACA}
            \label{fig:fs_10_aca_learning}
        \end{subfigure}
        \begin{subfigure}[t]{0.24\textwidth}
            \includegraphics[width=\textwidth]{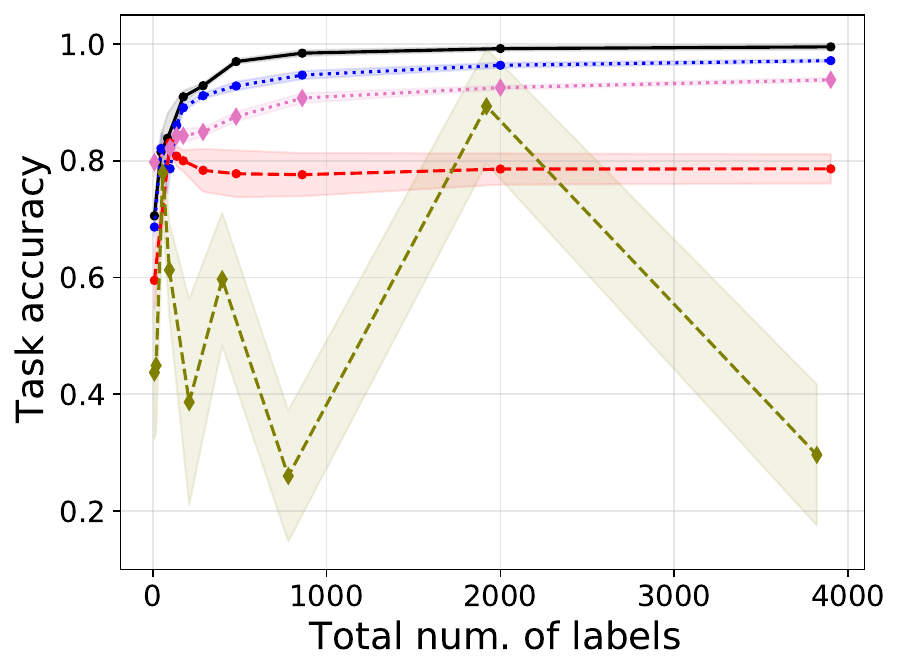}
            \caption{Plant Hitting Sets}
            \label{fig:plant_hs_learning}
        \end{subfigure}
        \caption{Task accuracy when the symbolic rules are learned.}
        \label{fig:learning_results}
\end{figure*}

The results are shown in Figures \ref{fig:addition_reasoning}-\ref{fig:e9p_reasoning} when the symbolic rules are given, and Figures \ref{fig:addition_learning}-\ref{fig:e9p_learning} when the rules are learned. \method{} outperforms all baselines, and learns the correct rules (see Supp. Material). In both cases, BLIP is fine-tuned to a mean image accuracy of 0.9216 with 100 labels, increasing to 0.9753 with 1000 labels. This is within 2.34\% of the state-of-the-art result of 0.9987 \cite{byerly2021no}, whilst using only 1.67\% of the 60,000 labels available. \method{} is able to learn the correct rules in all 5 repeats when BLIP is fine-tuned using only 10 image labels. To do so, the symbolic learner requires 50 task labels for the Addition task, and 20 for the E9P task. Note that the \method{} curves in Figures \ref{fig:addition_reasoning}-\ref{fig:e9p_reasoning} are shifted to the left because no task examples are required as the rules are given. All baselines use the same total number of examples as \method{} does when learning the rules (Figure \ref{fig:learning_results}).

In terms of the baselines in Figure \ref{fig:reasoning_results}, all methods except SLASH perform better on the E9P task than Addition. This is expected as the downstream label is more informative in E9P: There is a reduced number of possible digit choices for certain examples, which gives a stronger signal for training the perception model. \method{} is unaffected by this, given it is fine-tuned directly with answers generated from image labels. With fewer labels, the end-to-end variant of Embed2Sym struggles to achieve a good clustering in the embedding space of the perception model in Addition, again because it's more difficult to distinguish the individual digits in these examples. In Figure \ref{fig:learning_results}, a similar pattern occurs, with the notable exception of BLIP+GPT4. It performs well in Addition, but fails to generate the correct rules in the E9P task. The E9P task is less common than Addition, and it is unlikely GPT4 has seen many examples of this during pre-training. The fully neural ViT+ReasoningNet struggles in such low-label settings, whereas the \ac{nesy} approaches perform better. NSIL has to explore more of the search space in Addition before converging to the correct rules, and FF-NSL follows a similar curve to \method{}, but has lower performance due to an inferior perception component.

\subsection{Follow Suit}
The Follow Suit domain requires learning the rules of a trick-based playing card game, called Follow Suit Winner. Each player plays a card from a standard deck of 52 cards, and the winner is the player with the highest ranked card with the same suit as the first player. This domain involves a large amount of choice for the symbolic features, and therefore presents a challenge for end-to-end methods. To investigate this, we introduce a new 10-player variant, which significantly increases the number of choices compared to the 4-player variant. Assuming a single deck of cards, the 4-player variant has $\sim6.5$m choices for the symbolic features, whereas the 10-player variant has $\sim5.74\times 10^{16}$ choices. The search space for the symbolic learner follows \cite{cunnington2023ffnsl} and includes the relations \asp{higher\_rank} of arity 2 which denotes a certain player having a higher ranked card than another player, and the `!=' relation to denote two suit values are not equal. The search space also includes an instruction to perform predicate invention, to enable the definition of a winning player to be learned. 

For images, we use the \textit{Standard} and \textit{Adversarial Captain America (ACA)} decks from \cite{cunnington2023ffnsl}, where the ACA deck contains card images from a Captain America card deck placed on a background of standard card images. BLIP is fine-tuned with the questions ``\textit{which rank is this?}", and ``\textit{which suit is this?}". 
Image accuracies are reported over a test set of 1040 card images, and task accuracies are reported over held-out test sets of 1000 examples.

The results for the most challenging 10-player ACA variant are shown in Figures \ref{fig:fs_10_aca_reasoning} and \ref{fig:fs_10_aca_learning} for reasoning and learning respectively. The results for the other variants are shown in the Supp. Material. Note that all of the methods in Figure \ref{fig:fs_10_aca_learning} have low standard error across 5 repeats. In both reasoning and learning, \method{} achieves perfect performance on all variants, and significantly outperforms the baselines. When the symbolic rules are given, NeurASP, SLASH, and BLIP+ABL all reach the 24-hour time limit, even with the 4-player standard variant. NeurASP struggles with the large number of choices, and we had to limit the number of answer sets calculated during each example to 5000, so it is not able to update gradients based on all the answer sets. We also had to limit NeurASP to 10 epochs. 
SLASH became stuck in the ASP grounding stage, and therefore we are unable to obtain any results. For BLIP+ABL, the 10-player variant takes longer because there are more images in each example to calculate pseudo-labels for. Embed2Sym* is able to learn a good playing card clustering in the embedding space, but the ASP Optimisation used to label the clusters struggles, and we had to limit the \ac{asp} solver to return the best cluster labelling discovered within a 60 second timeout. The end-to-end variant of Embed2Sym also struggles with cluster labelling, and is unable to learn a good clustering in the embedding space. See the Supp. Material for the embedding space plots.

When learning the rules, FF-NSL is the next-best baseline, but requires more labels to fine-tune the perception component to the same level of accuracy. As NSIL is based upon NeurASP, it suffers similar timeout issues. The fully neural VIT+ReasoningNet struggles in such low-label settings. BLIP+GPT4 and BLIP Single Image perform well in the 4-player variants, with BLIP Single Image achieving perfect accuracy given enough labels in the 4-player standard case. However, both of these approaches fail to solve the 10-player variant. 
For NeSyGPT, BLIP is fine-tuned to an image accuracy of 0.9832 and 0.9988 for standard and ACA decks respectively, with only 52 image labels (single-shot). This increases to 1.0 in all 5 repeats for both standard and ACA variants, given 520 and 2600 labels respectively. 
The correct rules are learned with only 10 task labels, in all variants.

\subsection{Plant Hitting Sets}
The goal is to learn a variation of the Hitting Set problem \cite{karp1972reducibility} from images of plant leaves of various crops \cite{singh2020plantdoc}. Let us assume a collection of crop fields, where each field contains a set of crops that can be diseased or healthy. In this context, a ``hitting set" represents a set of diseased crops that intersects with each crop field. This enables a prioritisation in terms of which diseases should be eradicated first, as one may wish to eradicate diseases that occur in each crop field. The size of the hitting set can be viewed as the budget for how many crops are able to be eradicated.  
The standard rules of the Hitting Set problem apply, 
and we also require the constraint that no healthy crops should appear in the hitting set, since we are not interested in eradicating those. To solve this task, complex rules are required involving constraints, choice rules, and, to generate \textit{all} the hitting sets with a certain budget, the solution requires multiple models.

The input consists of a collection of plant disease images arranged into various subsets. We generate the examples randomly, where each example contains at most 5 subsets, and each subset contains at most 5 elements. In the plant disease image dataset \cite{singh2020plantdoc}, each element can take an integer value in $\{1..38\}$ to denote a particular plant species and disease. Note we remove the ``background" class. We assume the budget is 2, and the label is binary, indicating whether the collection contains a hitting set with no healthy images of size $\leq 2$. The search space for the symbolic learner contains the relation `!=' and the relation $\asp{hs}$, which defines an element of the hitting set. The budget 2, as well as the integer values which correspond to healthy plant images, are given as background knowledge. To fine-tune BLIP, we use the questions ``\textit{which plant species is this?}", and ``\textit{which disease is this?}". If the plant image is not diseased, we use the answer ``\textit{healthy}". Image accuracies are reported over a test set of 5000 plant images, and task accuracies are reported over held-out test sets of 1000 examples.

The results are shown in Figures \ref{fig:plant_hs_reasoning} and \ref{fig:plant_hs_learning}, for reasoning and learning respectively. With $>10$ labels, \method{} is able to match or out-perform all the baseline methods. BLIP is fine-tuned to an accuracy of 0.8826 with 380 image labels, and 0.9871 with 3800 labels. The correct rules are learned using the BLIP model fine-tuned with 380 image labels, when 100 examples are given to the symbolic learner.  

In terms of the baselines, when the rules are given, NeurASP and BLIP+ABL suffer similar timeout issues to Follow Suit, and the implementation of SLASH does not support variable length sequences in the examples, so results can't be obtained. Both variants of Embed2Sym are fairly constant, regardless of the number of labels. Inspecting the perception model's embedding space, it appears the pre-trained vision transformer model is able to return a reasonably good image clustering without any fine-tuning. However, labelling the clusters accurately is very difficult, since there are many possible choices of symbolic features that satisfy the downstream label (see the Supp. Material for details). When learning the rules, FF-NSL performs similarly to \method{}, as does BLIP Single Image. The ViT+ReasoningNet is unable to solve the task, although when the reasoning network (a Recurrent Transformer) is given 5000 downstream examples (50$\times$ more than shown in Figure \ref{fig:plant_hs_learning}), the task accuracy approaches 90\%. The results for BLIP+GPT4 are very variable, and somewhat random. We note that this method requires extensive prompt tuning, although is able to learn the correct rules on some repeats. We had to limit the number of downstream examples to a maximum of 20 as otherwise the token limit in the prompt was exceeded. 

\subsection{CLEVR-Hans}
To demonstrate \method{} can solve tasks that require detecting multiple objects within an image, we present results on the CLEVR-Hans3 dataset \cite{stammer2021right}. The input is a single image containing up to 10 objects, and the downstream label is one of three classes, which groups images with certain objects and properties together. To solve this task, BLIP must detect the properties of the various objects, namely, their \textit{size}, \textit{material}, \textit{color}, and \textit{shape}. The learned rules then classify a set of objects and their properties. The search space for the symbolic learner contains all possible properties and their values. BLIP is fine-tuned using the question ``\textit{what does the image contain?}" and the answer contains the list of object properties, e.g., ``\textit{The image contains a large gray cube made of metal, a large yellow cylinder made of rubber, ...}". Image and task accuracies are reported over the CLEVR-Hans test set, where we count an image prediction as correct if all ground truth objects and their properties are correctly identified. For training, we fine-tune BLIP firstly using the 70k-example CLEVR dataset, as performed in \cite{shindo2023alpha}, and secondly, using the 9000-example CLEVR-Hans training set. For rule learning, we use 500 and 250 task examples. In all cases, we run 5 repeats with different randomly generated seeds. For $\alpha$ILP, we use the 5 seeds specified in their implementation.




\begin{table}[t]
\centering
\resizebox{1\linewidth}{!}{%
\begin{tabular}{@{}lrrr@{}}
\toprule
                          & Num Im. & Task Acc. (250 Task Ex.) & Task Acc. (500 Task Ex.)\\ \midrule

\multirow{2}{*}{$\alpha$ILP} & 70k    & 0.9505 (0.0024) & 0.9752 (0.0081)\\
 & 9k     & 0.4992 (0.0116) & 0.4808 (0.0160)\\ \midrule
\multirow{2}{*}{NeSyGPT}  & 70k    & \textbf{0.9853} (0.0012) & \textbf{0.9853} (0.0012)\\
 & 9k     & \textbf{0.8629} (0.0090) & \textbf{0.8691} (0.0085)\\ \bottomrule

\end{tabular}
}
\caption{CLEVR-Hans Results.}
\label{tab:clevr_hans}
\end{table}

The results are shown in Table \ref{tab:clevr_hans}. \method{} outperforms $\alpha$ILP in all cases, achieving state-of-the-art performance using 50\% of the downstream labels. BLIP is fine-tuned to an image accuracy of 0.410 and 0.934, and the average rule accuracy with 250 task examples is 0.831 and 1.0, for the 9000 and 70k cases respectively. With 500 task examples, the average rule accuracy increases to 0.833 in the 9000 case.

\section{Conclusion}
In this paper, we have introduced a new architecture called \method{}, that integrates a vision-language foundation model with symbolic learning and reasoning in \ac{asp}. Our evaluation demonstrates \method{} provides many benefits over existing \ac{asp}-based \ac{nesy} systems, including strong performance, reduced labelling, advanced perception, and symbolic learning. We have also demonstrated that \acp{llm} can be used to reduce the manual engineering effort, generating questions and answers to fine-tune the perception model, and also the programmatic interface between the neural and symbolic components. Future work could explore generating the search space for the symbolic learner using a \ac{llm}, and developing \ac{vqa} fine-tuning in more advanced models, which could then be integrated into the \method{} architecture.

\appendix

\section*{Ethical Statement}
Our approach relies upon naturally interpretable logic programs for the symbolic component, which helps to increase transparency of what has been learned, compared to a fully neural system. Also, these programs can be manually adjusted if required, and provide formal guarantees of their behaviour when deployed to production. However, our approach also relies upon the BLIP vision-language foundation model, which is difficult to interpret and assert strong claims over its future behaviour. Through our approach of fine-tuning with questions and answers generated from image labels, we hope to reduce the risk of hallucinations in the specific tasks and settings our system is deployed. Also, the modular nature of our system enables each component to be adjusted and updated if required. The usual cautions and monitoring should apply if our approach is used in a production setting, particularly if image inputs are observed that are significantly outside the distribution in which BLIP has been fine-tuned.


\bibliographystyle{named}
\bibliography{ijcai24}

\clearpage
\appendix
\section{Appendix}
\subsection{Reducing the Engineering using a LLM}
We evaluate GPT-4 \cite{openai2023gpt4} when generating questions and answers for fine-tuning BLIP, and when generating the training examples for the symbolic learner. In both cases, we prompt GPT-4 with instructions and an example.

\textbf{QA Generator}. We use a base prompt, followed by an instruction for the given task. In this case, we require the user to input a natural language description of what the image represents, the space of possible image features, and a description of the downstream task. Note that we don't specify the specific rules of the task in the description. The base prompt used across all tasks is shown in Listing \ref{lst:qa_base}, and the specific instructions used for the MNIST Arithmetic, Follow Suit, and Plant Hitting Sets tasks are shown in Listings \ref{lst:arith_qa}, \ref{lst:fs_qa}, and \ref{lst:phs_qa} respectively.

\begin{lstlisting}[caption={Base prompt used across all tasks for generating QA for fine-tuning BLIP.}, label={lst:qa_base}]
You are required to generate a set of questions and 
answers to extract symbolic features from an image. You 
are given a description of the type of object the image 
represents, a description of the downstream task to be 
performed based on extracted features, and a list of 
ground-truth labels which denote the possible object 
instances. Return your response as a JSON list of 
questions and answers for each label. Try to use the exact 
same question(s) for all object instances, and ensure 
the answers are unique across each instance such that a 
classification can be performed. Under no circumstances 
should you return a response where two different labels 
have the same answers to the same question(s). Simple 
questions that distinguish between all instances are 
preferred over detailed questions on specific features. 
Use as few questions as possible, and eliminate redundant 
questions. Here is an example: 

Description: The image represents a vehicle. 
Downstream task description: The task is to learn rules 
    based on the type of vehicle. 
Labels: [car, ship, bicycle]. 

Output: 

```json
{
  "car": [
    {"question": "Does the vehicle have wheels?", "answer": "yes"},
    {"question": "Does the vehicle drive on a road?", "answer": "yes"},
    {"question": "Does the vehicle carry passengers?", "answer": "yes"}
  ],
  "ship": [
    {"question": "Does the vehicle have wheels?", "answer": "no"},
    {"question": "Does the vehicle drive on a road?", "answer": "no"},
    {"question": "Does the vehicle carry passengers?", "answer": "yes"}
  ],
  "bicycle": [
    {"question": "Does the vehicle have wheels?", "answer": "yes"},
    {"question": "Does the vehicle drive on a road?", "answer": "yes"},
    {"question": "Does the vehicle carry passengers?", "answer": "no"}
  ]
}```
\end{lstlisting}

\begin{lstlisting}[caption={Instruction used for MNIST Arithmetic QA generation.}, label={lst:arith_qa}]
Generate a set of questions and answers as follows. 
Description: The image represents a handwritten digit 0-9. 
Downstream task description: Perform arithmetic over 
    digits. 
Labels: [0,1,2,3,4,5,6,7,8,9]. 

Output: 
\end{lstlisting}

\begin{lstlisting}[caption={Instruction used for Follow Suit QA generation.}, label={lst:fs_qa}]
Generate a set of questions and answers as follows. 
Description: The image represents a playing card from a 
    standard deck of 52 cards.
Downstream task description: Learn the rules of a card 
    game. 
Labels: ["two of hearts", "two of diamonds", "two of 
    spades", "two of clubs", ..., "ace of clubs"].

Do not truncate the response.

Output: 
\end{lstlisting}

\begin{lstlisting}[caption={Instruction used for Plant Hitting Sets QA generation.}, label={lst:phs_qa}]
Generate a set of questions and answers as follows. 
Description: The image represents a plant leaf. The plant 
    leaf is from a certain species and may be diseased or 
    healthy. 
Downstream task description: Learn rules over groups of 
    plant species and diseases. 
Labels: ["apple scab", "apple black rot", ..., "tomato 
    healthy"].

Output: 
\end{lstlisting}

\begin{table*}[t]
\centering
\resizebox{\textwidth}{!}{%
\begin{tabular}{@{}lrrrrrrrr@{}}
\cmidrule(l){2-8}
                      & \multicolumn{2}{c}{\textbf{MNIST Arithmetic}}                             & \multicolumn{2}{c}{\textbf{Follow Suit 4}}                               & \multicolumn{2}{c}{\textbf{Follow Suit 10}}                              & \multicolumn{1}{c}{\textbf{Plant HS}} \\
                      
                      & \multicolumn{2}{c}{100+100 labels}                             & \multicolumn{2}{c}{52+100 labels}                               & \multicolumn{2}{c}{52+100 labels}                              & \multicolumn{1}{c}{760+100 labels}
                      \\ \cmidrule(l){2-8}
                      & \multicolumn{1}{c}{\textbf{Sum}} & \multicolumn{1}{c}{\textbf{E9P}} & \multicolumn{1}{c}{\textbf{Standard}} & \multicolumn{1}{c}{\textbf{ACA}} & \multicolumn{1}{c}{\textbf{Standard}} & \multicolumn{1}{c}{\textbf{ACA}} & \multicolumn{1}{c}{}                  \\ \midrule
Manual QA + Manual Sym. Ex. & \textbf{0.8492} (0.01) & \textbf{0.9004} (0.00) & \textbf{0.9958} (0.00) & \textbf{0.9994} (0.00) & \textbf{0.9968} (0.00) & \textbf{1.0000} (0.00) & \textbf{0.9846} (0.00) \\ 
LLM QA + Manual Sym. Ex. & 0.6954 (0.16) & 0.7302 (0.17) & 0.9928 (0.00) & 0.9984 (0.00) & 0.9892 (0.01) & 0.9974 (0.00) & 0.9734 (0.01) \\ 
Manual QA + LLM Sym. Ex. & 0.5158 (0.21) & 0.5486 (0.22) & 0.2938 (0.20) & 0.2938 (0.20) & 0.2936 (0.18) & 0.2936 (0.18) & 0.7874 (0.20) \\ 
LLM QA + LLM Sym. Ex. & 0.1634 (0.16) & 0.1746 (0.17) & 0.0500 (0.05) & 0.0500 (0.05) & 0.0908 (0.09) & 0.0912 (0.09) & 0.1974 (0.20) \\ 
\midrule
Manual QA + LLM Sym. Ex.$^*$ & \textbf{0.8492} (0.01) & \textbf{0.9004} (0.00) & 0.5986 (0.24) & 0.5994 (0.24) & \textbf{0.9968} (0.00) & \textbf{1.0000} (0.00) & - \\ 

LLM QA + LLM Sym. Ex.$^*$ & 0.6832 (0.17) & 0.7198 (0.18) & 0.8472 (0.15) & 0.8486 (0.15) & 0.7898 (0.20) & 0.7976 (0.20) & - \\ 

\end{tabular}
}
\caption{Task accuracy with manual vs LLM engineering. The number of labels is decomposed as ``num image labels + num task labels". * denotes manual fixes to simple errors.}
\label{tab:reduced_engineering}
\end{table*}

\textbf{Example Generator} Again, we use a base prompt followed by an instruction specific to each task. We require the user to input a description of the downstream task, with details on the features given, and the range of input and output values. The ASP background knowledge and mode declarations are also required, and we also input the full QA JSON that is either manually specified, or LLM generated. The base prompt is shown in Listing \ref{lst:sym_ex_base}, and the instructions used for the MNIST Arithmetic, Follow Suit, and Plant Hitting Sets tasks are shown in Listings \ref{lst:arith_sym_ex}, \ref{lst:fs_sym_ex}, and \ref{lst:phs_sym_ex} respectively. In this paper, to avoid duplication, we omit the ASP background knowledge and mode declarations, as these are presented in Section \ref{sec:app:bk}.

\begin{lstlisting}[caption={Base prompt used across all tasks for generating symbolic training examples.}, label={lst:sym_ex_base}]
The task is to generate a python function called 
"create_wcdpi" that takes as input a list of image 
features in the form of questions and answers, together 
with a label and a unique identifier, and returns a 
training example for a symbolic learning system based on 
Answer Set Programming (ASP). The symbolic learning system 
learns a set of logical rules in the form of an ASP 
program to solve a downstream task. A training example
can be positive or negative, and it consists of the unique
identifier, an integer weight, a set of inclusions,
exclusions, and a set of contextual facts. The inclusion
set can represent the label that should be output by the 
learned rules, given the contextual facts. The exclusion
set can represent the labels that should *not* be output
given the contextual facts. For binary classification 
tasks, the label can be represented with a positive (1) or 
negative (0) example, and the inclusion and exclusion sets
can remain empty. Note the atoms within the inclusion
and exclusion sets, as well as the atoms in the contextual
facts are specified in the language of ASP. To generate
a symbolic training example, the WCDPI python class can be
used. This is instantiated as follows: new_example =
WCDPI(f"id{example_identifier}", positive=True, weight=10,
inclusion=[], exclusion=[], context=[]). The "example_id"
attribute is the unique identifier, the boolean "positive" 
attribute indicates whether the example is positive or 
negative, the weight is the integer weight and should be 
set to 10, and the remaining attributes correspond to the 
inclusion, exclusion, and context sets respectively. To 
help generate the correct python function, you are given a 
description of the downstream task, a complete set of 
possible questions and answers for each type of image, as 
well as an ASP background knowledge specification 
related to the downstream task. Finally, you are also 
given a set of mode declarations, which the 
symbolic learner uses to construct a search space for 
possible rules. The mode declarations define what should 
be in the head and the body of the learned rules, using 
#modeh, and #modeb statements respectively. The background 
knowledge and mode declarations give clues as to the 
format of the atoms in the inclusion and exclusion sets, 
as well as the contextual facts.

Here are some examples:

Example 1:
Description: The downstream task is to learn a set of 
    rules define the validity of 4x4 Sudoku grids. Image 
    features are given related to the digits apparent in 
    each cell. The digit values are in the range 1-4, and
    the output is a binary classification. Note that it 
    is important to associate the value of each digit to 
    a particular cell on the Sudoku grid.

Complete set of questions and answers for each type of 
    image: {"1": [{"question": "what is the digit value?", 
    "answer": "1"}], "2": [{"question": "what is the digit 
    value?", "answer": "2"}], "3": [{"question": "what 
    is the digit value?", "answer": "3"}], "4": [{
    "question": "what is the digit value?", "answer": 
    "4"}]}

ASP background knowledge for the downstream task: 
value("1, 1", X) :- value(1, 1, X).
value("1, 2", X) :- value(1, 2, X).
value("1, 3", X) :- value(1, 3, X).
value("1, 4", X) :- value(1, 4, X).
value("2, 1", X) :- value(2, 1, X).
value("2, 2", X) :- value(2, 2, X).
value("2, 3", X) :- value(2, 3, X).
value("2, 4", X) :- value(2, 4, X).
... etc.

% Columns
col("1, 1", 1).
col("1, 2", 2).
col("1, 3", 3).
col("1, 4", 4).
col("2, 1", 1).
col("2, 2", 2).
col("2, 3", 3).
col("2, 4", 4).
... etc.

% Rows
row("1, 1", 1).
row("1, 2", 1).
row("1, 3", 1).
row("1, 4", 1).
row("2, 1", 2).
row("2, 2", 2).
row("2, 3", 2).
row("2, 4", 2).
... etc.


% Blocks
block("1, 1", 1).
block("1, 2", 1).
block("2, 1", 1).
block("2, 2", 1).
... etc.

Mode declarations for the symbolic learner: 
#modeh(invalid).
#modeb(value(var(cell), var(num))).
#modeb(row(var(cell), var(row))).
#modeb(not row(var(cell), var(row))).
#modeb(col(var(cell), var(col))).
#modeb(not col(var(cell), var(col))).
#modeb(block(var(cell), var(block))).
#modeb(not block(var(cell), var(block))).
#modeb(neq(var(cell), var(cell))).
neq(X, Y) :- cell(X), cell(Y), X != Y.

#maxv(4).
num(1..4).
row(1..4).
col(1..4).
block(1..4).
cell(C) :- value(C, _).

#bias("penalty(1, head).").
#bias("penalty(1, body(X)) :- in_body(X).").

Python function:

```python
def create_wcdpi(example_identifier: str, image_features: list, label: str):    
    # Decide whether to create a positive or negative example
    if label == '1':
        positive = True
    else:
        positive = False
    
    # Create contextual facts
    # Assume image features are passed in order of increasing row and column
    ctx = []
    GRID_SIZE = 4
    current_idx = 0
    for row in range(1, GRID_SIZE+1):
        for col in range(1, GRID_SIZE+1):
            ctx.append(f'value({row},{col},{image_features[current_idx]["what is the digit value?"]}).')
            current_idx += 1
    
    # Create WCDPI
    new_example = WCDPI(ex_id=f"id{example_identifier}", positive=positive, weight=10, inclusion=[], exclusion=[], context=ctx)
    
    return new_example
```

Example 2:
Description: The downstream task is to learn a set of rules that define the features of various animals. Image features are given related to a single animal (one image). Only one animal is present in each image.
Complete set of questions and answers for each type of image: {"fish": [{"question": "does it have hair?", "answer": "no"}, {"question": "does it have fins?", "answer": "yes"}, {"question": "does it lay eggs?", "answer": "yes"}, {"question": "does it have feathers?", "answer": "no"}, {"question": "does it have legs?", "answer": "no"}], "bird": [{"question": "does it have hair?", "answer": "no"}, {"question": "does it have fins?", "answer": "no"}, {"question": "does it lay eggs?", "answer": "yes"}, {"question": "does it have feathers?", "answer": "yes"}, {"question": "does it have legs?", "answer": "yes"}], "mammal": [{"question": "does it have hair?", "answer": "yes"}, {"question": "does it have fins?", "answer": "no"}, {"question": "does it lay eggs?", "answer": "no"}, {"question": "does it have feathers?", "answer": "no"}, {"question": "does it have legs?", "answer": "yes"}]}
ASP background knowledge for the downstream task: 
:- class(X), class(Y), X < Y.

Mode declarations for the symbolic learner: 
#maxv(1).
#modeh(class(const(class))).
#modeb(1, hair(const(hair))).
#modeb(1, feathers(const(feathers))).
#modeb(1, eggs(const(eggs))).
#modeb(1, fins(const(fins))).
#modeb(1, legs(const(legs))).

Python function:

```python
def create_wcdpi(example_identifier: str, image_features: list, label: str):    
    # Inclusion set
    inclusion_set = [f'label({label})']
    
    # Exclusion set - all other classes
    classes = ['fish', 'bird', 'mammal']
    exclusion_set = [f'label({l})' for l in classes if l != label]

    # Create contextual facts
    # Single image so extract first image features
    first_image = image_features[0]
    ctx = [f'hair({first_image["does it have hair?"]}).', f'fins({first_image["does it have fins?"]}).', 
    f'feathers({first_image["does it have feathers?"]}).', f'eggs({first_image["does it lay eggs?"]}).',
    f'legs({first_image["does it have legs?"]}).']

    # Create WCDPI
    new_example = WCDPI(ex_id=f"id{example_identifier}", positive=True, weight=10, inclusion=inclusion_set, exclusion=exclusion_set, context=ctx)

    return new_example
```
\end{lstlisting}

\begin{lstlisting}[caption={Instruction used for MNIST Arithmetic symbolic example generation.}, label={lst:arith_sym_ex}]
Goal:
Generate a Python function for the following task:
Description: The downstream task is to learn a set of 
    rules that perform arithmetic over two numeric digits. 
    Image features are given related to the two digits, and 
    the digit values are in the range 0-9. The output is in 
    the range 0-18.

Complete set of questions and answers for each type of 
    image: 
    {'0': [{'question': 'which digit is this?', 'answer': 
    '0'}], 
    ...,
    '9': [{'question': 'which digit is this?', 'answer': 
    '9'}]}

<<Insert ASP background knowledge and mode declarations>>

Python function: 
\end{lstlisting}

\begin{lstlisting}[caption={Instruction used for Follow Suit symbolic example generation.}, label={lst:fs_sym_ex}]
Goal:
Generate a Python function for the following task:
Description: The downstream task is to learn a set of rules of a playing card game with 4 players. Image features are given related to 4 playing card images, which correspond to a card played by each player of the game. The cards take values from a standard deck of 52 playing cards, and the output indicates which player wins the game.

Complete set of questions and answers for each type of image: {'2h': [{'question': 'which color are the symbols?', 'answer': 'red'}, {'question': 'which symbol is on the card?', 'answer': 'heart'}, {'question': 'which suit is the playing card?', 'answer': 'hearts'}, {'question': 'which rank is the playing card?', 'answer': '2'}, {'question': 'is it a number card or face card?', 'answer': 'number'}], 
...,
'ac': [{'question': 'which color are the symbols?', 'answer': 'black'}, {'question': 'which symbol is on the card?', 'answer': 'club'}, {'question': 'which suit is the playing card?', 'answer': 'clubs'}, {'question': 'which rank is the playing card?', 'answer': 'ace'}, {'question': 'is it a number card or face card?', 'answer': 'face'}]}

<< Insert ASP background knowledge and mode declarations>> 

Python function: 
\end{lstlisting}

\begin{lstlisting}[caption={Instruction used for Plant Hitting Sets symbolic example generation.}, label={lst:phs_sym_ex}]
Goal:
Generate a Python function for the following task:
Description: The downstream task is to learn a set of rules over a collection of sets containing images related to plant disease. There are various plant diseases, and they are denoted by integers 1-38 inclusive. It is important to associate each image with a particular set identifier. The output is a binary classification. Assume a global variable "ss_element_assignments" is available, that is assigned a list that contains subset assignments for each image in an example. This list can be indexed by the example identifier to return a list of subset identifiers, one for each image. Also, assume a function "get_image_class" is available, that takes as input the key "{species}_{disease}", and returns the associated integer in the range 1-38.

Complete set of questions and answers for each type of image: {'1': [{'question': 'which plant species is this?', 'answer': 'apple'}, {'question': 'which disease is this?', 'answer': 'scab'}], 
...,
'38': [{'question': 'which plant species is this?', 'answer': 'tomato'}, {'question': 'which disease is this?', 'answer': 'healthy'}]}

<<Insert ASP background knowledge and mode declarations>>
  
Python function:
\end{lstlisting}

In order to evaluate the responses from the LLM, we perform a variety of experiments with all combinations of manual and LLM engineered components, and report the mean task accuracy, and standard error over 5 repeats. We use 5 different random seeds, and pass these to the LLM accordingly. All experiments are performed with the OpenAI API, and queries were made between 19th-30th December 2023. For the QA generation prompts, we use the ``gpt-4" model with 1000 max tokens for MNIST Arithmetic, and the ``gpt-4-1106-preview" model with 4096 max tokens for Follow Suit and Plant Hitting Sets. For the symbolic example prompts, we use the ``gpt-4" model with 750 max tokens in all tasks. Note on the Follow Suit task, we give only 50\% of the labels in the QA JSON, as otherwise the max token limit is exceeded.

The results are shown in Table \ref{tab:reduced_engineering}. Manually engineering everything achieves the best performance, although the LLM shows promise in generating questions and answers for fine-tuning BLIP in the Follow Suit and Plant Hitting Sets tasks (see \textit{LLM QA + Manual Sym.Ex}). In the MNIST Arithmetic tasks, it is difficult to decompose the MNIST images into a set of questions. On four repeats, GPT-4 returns the expected response of directly querying the value of the digit, whereas on one repeat it returns a set of questions based on the digit being even, and greater than 5. When GPT-4 directly generates the function to create symbolic examples, the accuracy is poor, although we show with simple syntactic fixes, the accuracy is able to approach that of manual engineering in the MNIST Arithmetic and the 10-player Follow Suit tasks. Simple errors include using a \asp{solution} predicate in the inclusion set instead of the \asp{label} predicate in the MNIST Arithmetic tasks, and using the rank values ``jack", ``queen", ``king", and ``ace", instead of single letters, as expected by the ASP background knowledge. In the Plant Hitting Sets case, the responses contained significant errors that couldn't easily be fixed manually. Full details of all responses, and errors made, are available in the code.\footnote{\url{https://www.github.com/TBC}} Requiring the user to apply simple fixes is a significantly reduced level of engineering compared to specifying the full components. In all but the Plant Hitting Sets domain, when using the LLM to generate both components, and assuming user fixes, the task accuracy achieved is reasonable. As LLMs are an active research area, we expect these results to improve as LLMs improve.

\subsection{Follow Suit Results}\label{app:fs_results}
\begin{figure*}[t]
    \centering
    \begin{subfigure}[t]{0.8\textwidth}
            \centering
            \includegraphics[width=1\linewidth]{figures/reasoning_legend.pdf}
        \end{subfigure}
        
        \begin{subfigure}[t]{0.24\textwidth}
            \includegraphics[width=\textwidth]{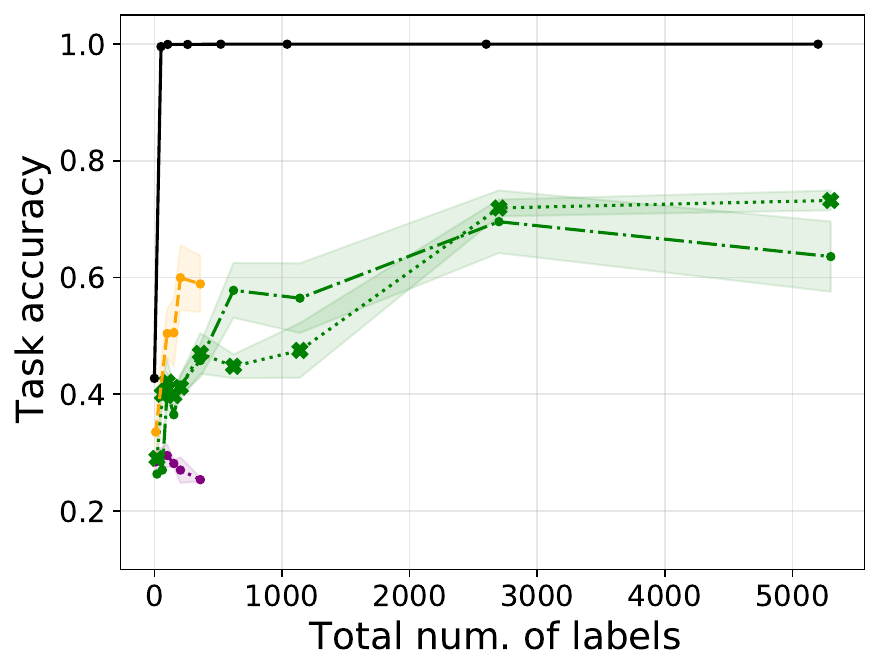}
            \caption{4 Standard}
            \label{fig:fs_4_standard_reasoning}
        \end{subfigure}
        \begin{subfigure}[t]{0.24\textwidth}
            \includegraphics[width=\textwidth]{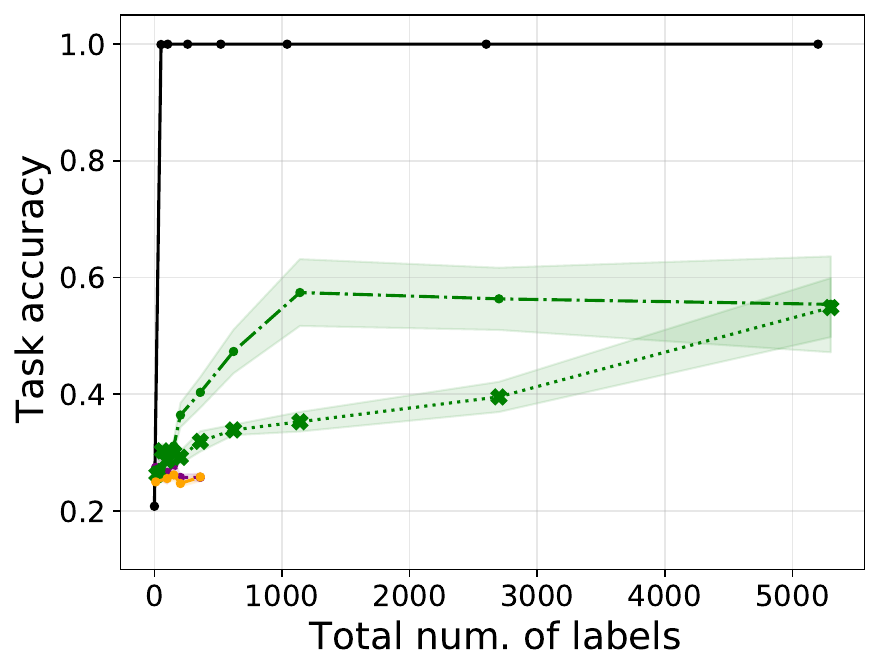}
            \caption{4 ACA}
            \label{fig:fs_4_aca_reasoning}
        \end{subfigure}
        \begin{subfigure}[t]{0.24\textwidth}
            \includegraphics[width=\textwidth]{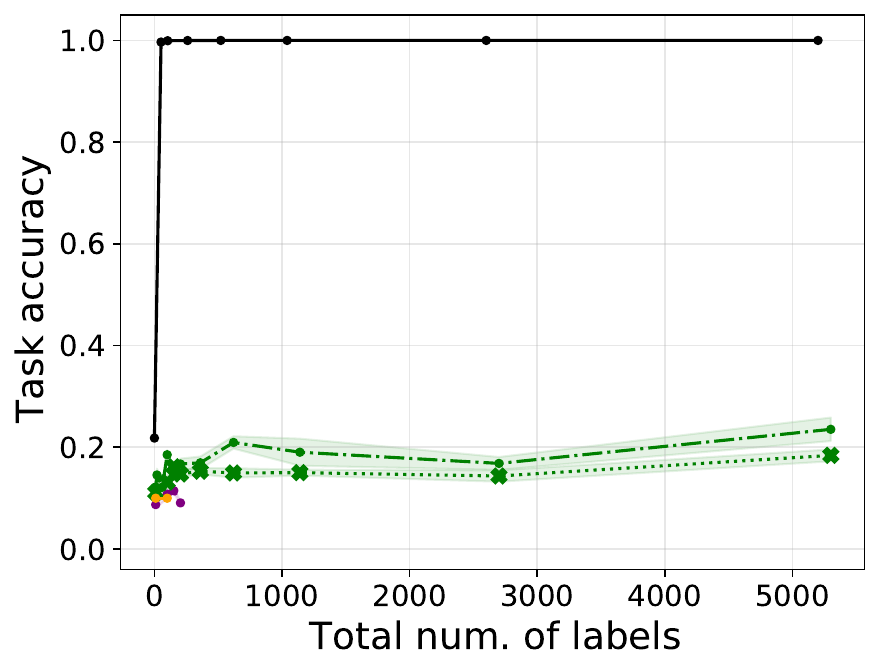}
            \caption{10 Standard}
            \label{fig:fs_10_standard_reasoning}
        \end{subfigure}
        \begin{subfigure}[t]{0.24\textwidth}
            \includegraphics[width=\textwidth]{figures/reasoning_follow_suit_10_aca.pdf}
            \caption{10 ACA}
            \label{fig:fs_10_aca_reasoning_app}
        \end{subfigure}
        \caption{Follow Suit task accuracy when the symbolic rules are given.}
        \label{fig:fs_reasoning_results}
\end{figure*}

\begin{figure*}[t]
    \centering
    \begin{subfigure}[t]{0.8\textwidth}
            \centering
            \includegraphics[width=1\linewidth]{figures/learning_legend.pdf}
        \end{subfigure}
        
         \begin{subfigure}[t]{0.24\textwidth}
            \includegraphics[width=\textwidth]{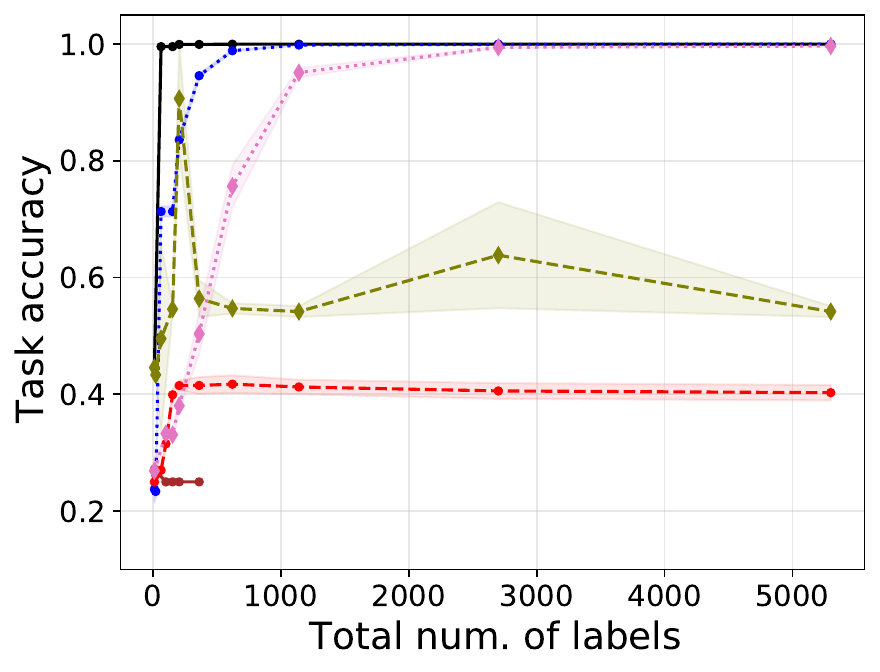}
            \caption{4 Standard}
            \label{fig:fs_4_standard_learning}
        \end{subfigure}
        \begin{subfigure}[t]{0.24\textwidth}
            \includegraphics[width=\textwidth]{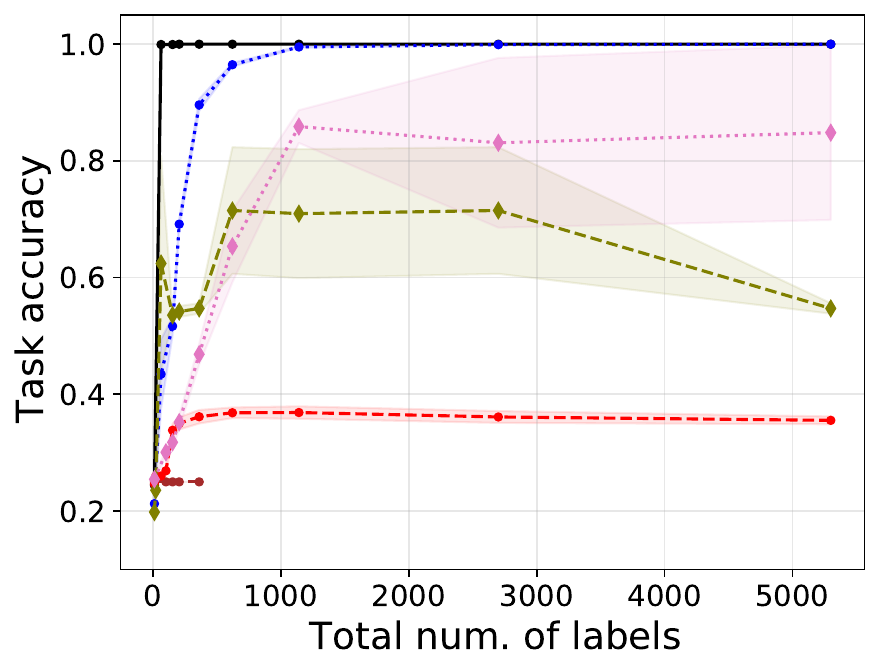}
            \caption{4 ACA}
            \label{fig:fs_4_aca_learning}
        \end{subfigure}
        \begin{subfigure}[t]{0.24\textwidth}
            \includegraphics[width=\textwidth]{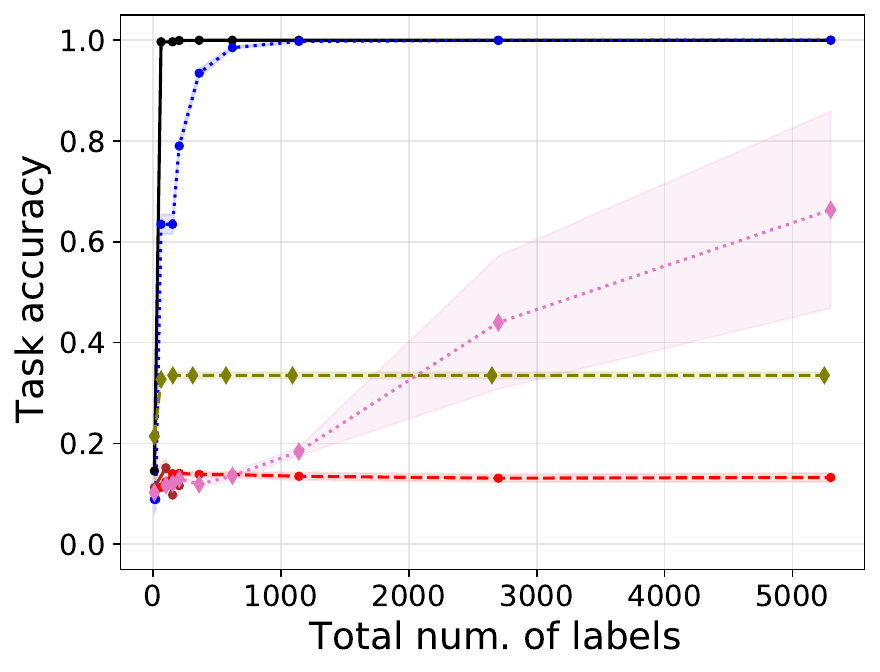}
            \caption{10 Standard}
            \label{fig:fs_10_standard_learning}
        \end{subfigure}
        \begin{subfigure}[t]{0.24\textwidth}
            \includegraphics[width=\textwidth]{figures/learning_follow_suit_10_aca.pdf}
            \caption{10 ACA}
            \label{fig:fs_10_aca_learning_app}
        \end{subfigure}
        \caption{Follow Suit task accuracy when the symbolic rules are learned.}
        \label{fig:fs_learning_results}
\end{figure*}

The Follow Suit results are presented in Figures \ref{fig:fs_reasoning_results} and \ref{fig:fs_learning_results} for reasoning and learning respectively. \method{} achieves perfect performance in all cases, and significantly outperforms the baselines, with the exception of FF-NSL, which is closer to our approach. As expected, all methods perform better on the 4-player standard and ACA variants, as the 10-player variant contains significantly more choice for the symbolic features. 

\subsection{Learned Rules}
The learned rules are shown in Figure \ref{fig:learned_rules}. For MNIST Arithmetic, the rules represent the Addition and E9P functions over two input digits, where the definitions of \asp{even} and \asp{plus\_nine} predicates are given in the background knowledge (see Section \ref{sec:app:bk}). For Follow Suit, the rules define whether a given player is a winner. \asp{winner(X)} is true if neither of the \asp{p1} rules hold. \asp{p1} is the invented predicate, and is true if the suit of the given player is different from the suit of Player 1, or if there is another player with the same suit as Player 1 with a higher ranked card than the given player. For Plant Hitting Sets, the first rule is a choice rule that generates all possible hitting sets, where \asp{V1} is the index and \asp{V2} is the element. The second rule defines whether a subset is ``hit", that is, if it contains an element in the hitting set. The remaining rules are constraints that ensure all subsets are hit, the hitting set must not contain different elements at a given index, and there are no ``healthy" elements in the hitting set. Finally, for CLEVR-Hans, the rules define the various object properties associated with each class, where \asp{V0} and \asp{V1} are object indices.

\begin{figure*}[t]
    \centering
    \begin{subfigure}[m]{0.32\textwidth}
        \begin{lstlisting}
<--Addition--!>
result(V0,V1,V2) :- V2 = V0 + V1.

<--E9P--!>
result(V0,V1,V2) :- even(V0),V2 = V1. 
result(V0,V1,V2) :- not even(V0), 
                           plus_nine(V1,V2).
    \end{lstlisting}
    \caption{MNIST Arithmetic}
    \label{fig:arithmetic_rules}
    \end{subfigure}
    \hspace{20pt}
    \begin{subfigure}[m]{0.38\textwidth}
        \begin{lstlisting}
winner(X) :- not p1(X), player(X).
p1(V1) :- V2 != V3, suit(V1,V2), suit(1,V3).
p1(V1) :- suit(V2,V3), suit(1,V3), 
              rank_higher(V2,V1).
    \end{lstlisting}    
        \caption{Follow Suit}
        \label{fig:fs_rules}
    \end{subfigure}
    \hspace{10pt}
    \begin{subfigure}[m]{0.34\textwidth}
        \begin{lstlisting}
0 {hs(V1,V2) } 1.
hit(V1) :- hs(V3,V2), ss_element(V1,V2).
:- ss_element(V1,V2), not hit(V1).
:- hs(V3,V1), hs(V3,V2), V1 != V2.
:- hs(V1,V2), healthy(V2).
    \end{lstlisting}
        \caption{Plant Hitting Sets}
        \label{fig:phs_rules}
        \end{subfigure}
        \hspace{20pt}
        \begin{subfigure}[m]{0.40\textwidth}
        \begin{lstlisting}
label(0) :- size(V0,large), shape(V0,cube), 
                size(V1,large), shape(V1,cylinder).
label(1) :- size(V0,small), material(V0,metal), 
                shape(V0,cube), size(V1,small), 
                shape(V1,sphere).
label(2) :- size(V0,large), color(V0,blue), 
                shape(V0,sphere), size(V1,small), 
                color(V1,yellow), shape(V1,sphere).
    \end{lstlisting}
        \caption{CLEVR-Hans}
        \label{fig:clevr_hans_rules}
        \end{subfigure}
        \caption{\method{} learned rules.}
        \label{fig:learned_rules}
\end{figure*}

\subsection{Domain Knowledge}\label{sec:app:bk}

\begin{lstlisting}[caption={MNIST Arithmetic domain knowledge.}, label={lst:arith_bk}]
% Background Knowledge
even(X) :- digit_type(X), X \ 2 = 0.
plus_nine(X1,X2) :- digit_type(X1), X2=9+X1.
:- digit(1,X0),digit(2,X1),label(Y1),label(Y2), Y1 != Y2.
label(Y) :- digit(1,X0), digit(2,X1), solution(X0,X1,Y).
num(0..18).
digit_type(0..9).

% Mode Declarations
#modeh(solution(var(digit_type),var(digit_type),
    var(num))).
#modeb(var(num) = var(digit_type)).
#modeb(var(num) = var(digit_type) + var(digit_type)).
#modeb(plus_nine(var(digit_type),var(num))).
#modeb(even(var(digit_type))).
#modeb(not even(var(digit_type))).
#maxv(3).
#bias("penalty(1, head(X)) :- in_head(X).").
#bias("penalty(1, body(X)) :- in_body(X).").
\end{lstlisting}

\begin{lstlisting}[caption={Follow Suit domain knowledge.}, label={lst:fs_bk}]
% Background Knowledge
suit(h). suit(s). suit(d). suit(c). rank(a). rank(2). 
rank(3). rank(4). rank(5). rank(6). rank(7). rank(8). 
rank(9). rank(10). rank(j). rank(q). rank(k). 
rank_value(2, 2). rank_value(3, 3). rank_value(4, 4). 
rank_value(5, 5). rank_value(6, 6). rank_value(7, 7). 
rank_value(8, 8). rank_value(9, 9). rank_value(10, 10). 
rank_value(j, 11). rank_value(q, 12). rank_value(k, 13). 
rank_value(a, 14). player(1..4).

% Highest rank
rank_higher(P1, P2) :- card(P1, R1, _), card(P2, R2, _), 
    rank_value(R1, V1), rank_value(R2, V2), V1 > V2.

% Link player's card to suit
suit(P1, S) :- card(P1, _, S).
        
% Mode Declarations
P(X) :- Q(X), identity(P, Q).
P(X) :- player(X), not Q(X), inverse(P, Q).
#modem(2, inverse(target/1, invented/1)).
#modem(2, identity(target/1, invented/1)).
#predicate(target, winner/1).
#predicate(invented, p1/1).
#modeh(p1(var(player))).
#modeb(1, var(suit) != var(suit)).
#modeb(1, suit(var(player), var(suit)), (positive)).
#modeb(1, suit(const(player), var(suit)), (positive)).
#modeb(1, rank_higher(var(player), var(player)), 
    (positive)).
#constant(player, 1).
#constant(player, 2).
#constant(player, 3).
#constant(player, 4).
\end{lstlisting}

\begin{lstlisting}[caption={Plant Hitting Sets domain knowledge.}, label={lst:phs_bk}]
% Background Knowledge
ss(1..5). hs_index(1..2). elt(1..38). healthy(4). 
healthy(5). healthy(7). healthy(11). healthy(15). 
healthy(18). healthy(20). healthy(23). healthy(24). 
healthy(25). healthy(28). healthy(38).
        
% Mode Declarations
#modeha(hs(var(hs_index), var(elt))).
#modeb(hs(var(hs_index), var(elt)),(positive)).
#modeb(var(elt) != var(elt)).
#modeb(ss_element(var(ss), var(elt)),(positive)).
#modeh(hit(var(ss))).
#modeb(hit(var(ss))).
#modeb(healthy(var(elt)), (positive)).
#bias(":- not lb(0), choice_rule.").
#bias(":- not ub(1), choice_rule.").
#bias(":- in_head(H1), in_head(H2), H1<H2.").
\end{lstlisting}

\begin{lstlisting}[caption={CLEVR-HANS domain knowledge.}, label={lst:clevr_hans_bk}]
% Background Knowledge
:- label(X), label(Y), X < Y. label_type(0..2). 
obj_id(1..10). s_type(small). s_type(large). 
m_type(rubber). m_type(metal). c_type(red). 
c_type(cyan). c_type(brown). c_type(yellow). 
c_type(blue). c_type(gray). c_type(purple). 
c_type(green). sh_type(cylinder). sh_type(sphere). 
sh_type(cube).

% Mode Declarations
#maxv(2).
#modeh(label(const(label_type))).
#modeb(size(var(obj_id), const(s_type))).
#modeb(material(var(obj_id), const(m_type))).
#modeb(color(var(obj_id), const(c_type))).
#modeb(shape(var(obj_id), const(sh_type))).
#bias("penalty(1, head(X)) :- in_head(X).").
#bias("penalty(1, body(X)) :- in_body(X).").
\end{lstlisting}

\subsection{Baseline Details}
In this section we present specific noteworthy details and analysis of the baseline methods.

\textbf{NeurASP}. On Follow Suit and Plant Hitting Sets tasks, we had to limit the number of epochs to 10, and the number of answer sets returned for each training example to 5000, in order to obtain results within the 24-hour time limit. 

\begin{figure}[H]
    \centering
    \begin{subfigure}[t]{0.45\linewidth}
            \centering
            \includegraphics[width=1\linewidth]{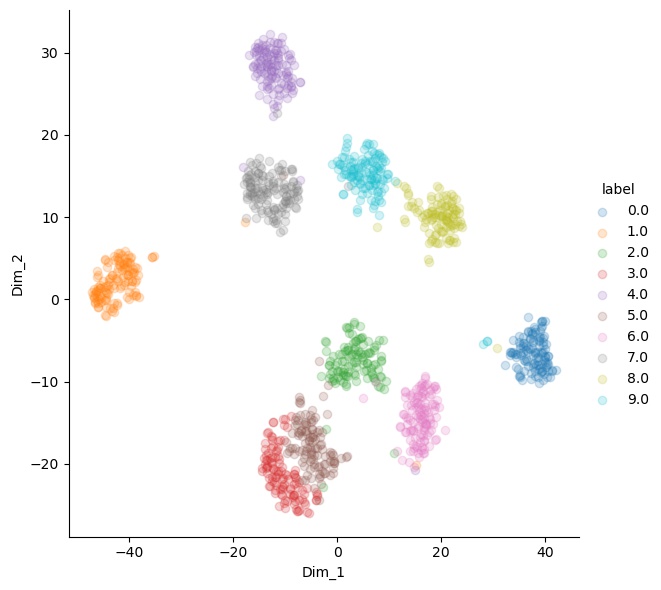}
            \caption{Trained with image labels}
        \end{subfigure}
        \begin{subfigure}[t]{0.45\linewidth}
            \includegraphics[width=\linewidth]{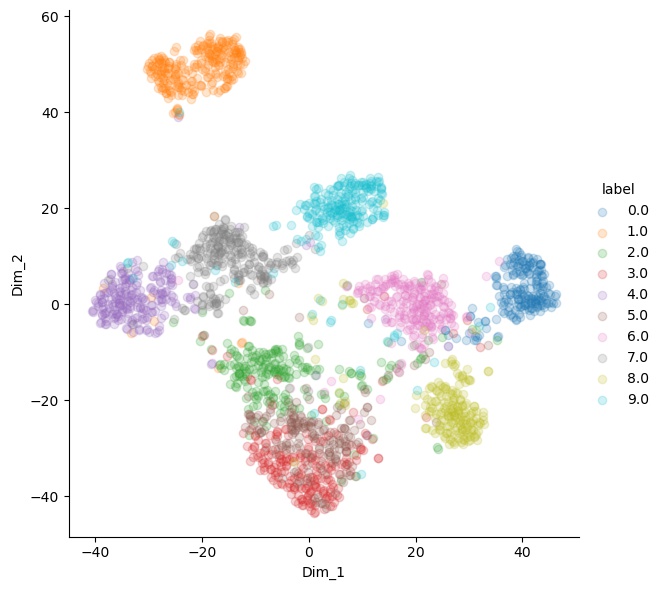}
            \caption{Trained end-to-end}
        \end{subfigure}
        \caption{Embed2Sym embedding spaces - Addition.}
        \label{fig:e2s_addition_embed}
\end{figure}

\begin{figure}[H]
    \centering
    \begin{subfigure}[t]{0.45\linewidth}
            \centering
            \includegraphics[width=1\linewidth]{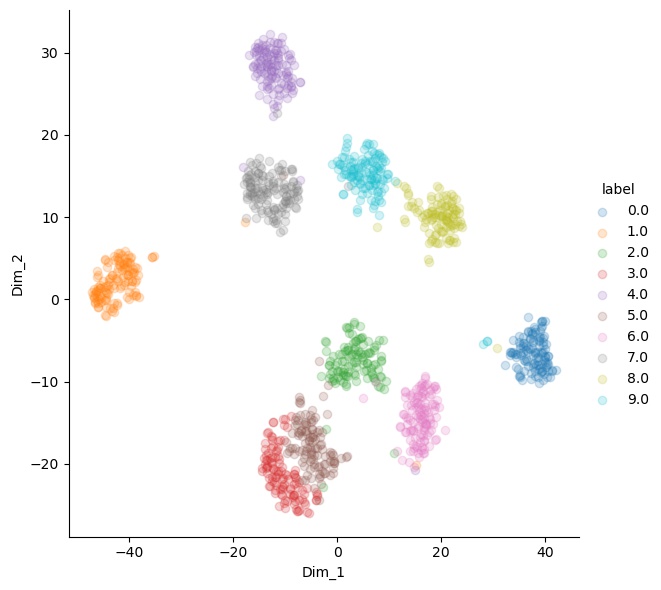}
            \caption{Trained with image labels}
        \end{subfigure}
        \begin{subfigure}[t]{0.45\linewidth}
            \includegraphics[width=\linewidth]{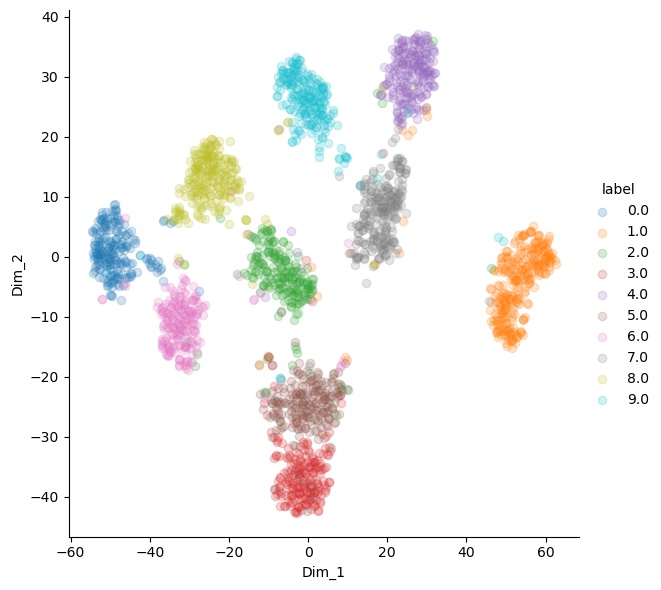}
            \caption{Trained end-to-end}
        \end{subfigure}
        \caption{Embed2Sym embedding spaces - E9P.}
        \label{fig:e2s_e9p_embed}
\end{figure}

\textbf{Embed2Sym}. In the MNIST Arithmetic domain, the variant of Embed2Sym trained with image features, Embed2Sym*, achieves a better clustering in the embedding space on the Addition task, compared to the end-to-end variant. This is because the downstream labels in the Addition task provide a weaker training signal than training directly with image labels. To illustrate this, the TSNE plots of the embedding spaces for the first repeat when the models are trained with 1100 labels are shown in Figure \ref{fig:e2s_addition_embed}. For reference, Figure \ref{fig:e2s_e9p_embed} shows the same plots but for the E9P task, where the clustering is similar between the two training strategies. This explains the performance difference in Figure \ref{fig:addition_reasoning}, and the similar performance in Figure \ref{fig:e9p_reasoning} of the two Embed2Sym variants.

\begin{figure}[H]
    \centering
    \begin{subfigure}[t]{0.45\linewidth}
            \centering
            \includegraphics[width=1\linewidth]{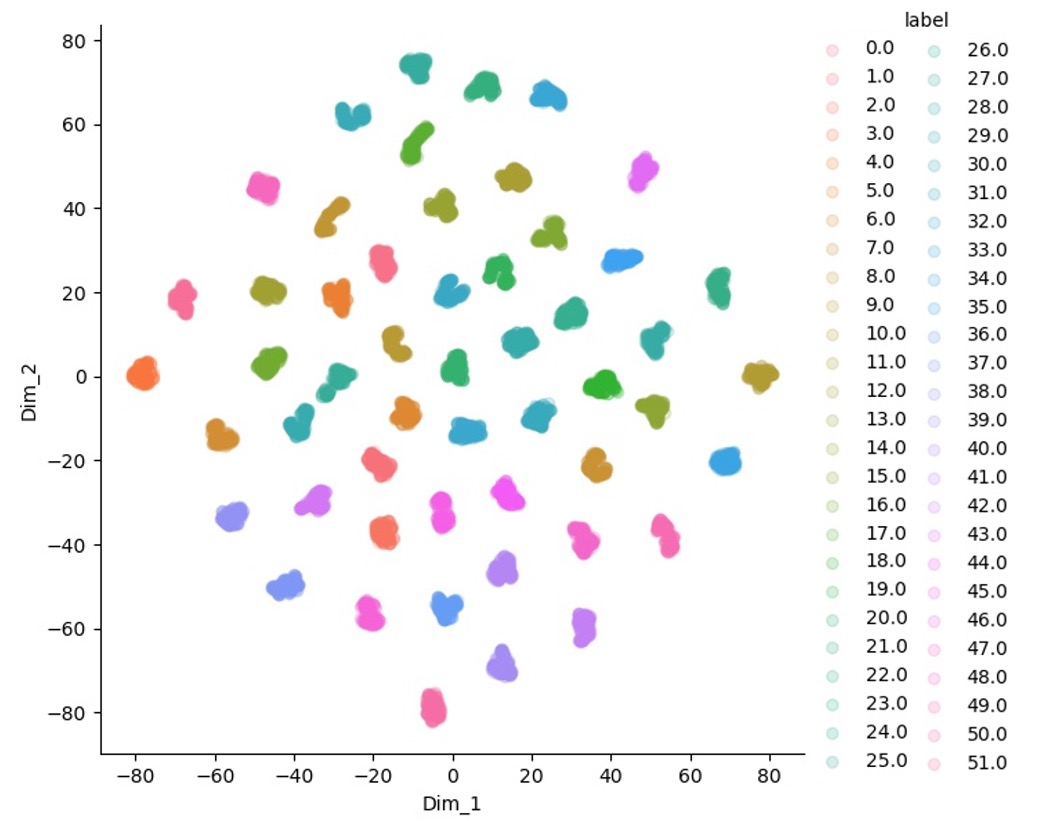}
            \caption{Trained with image labels}
        \end{subfigure}
        \begin{subfigure}[t]{0.45\linewidth}
            \includegraphics[width=\linewidth]{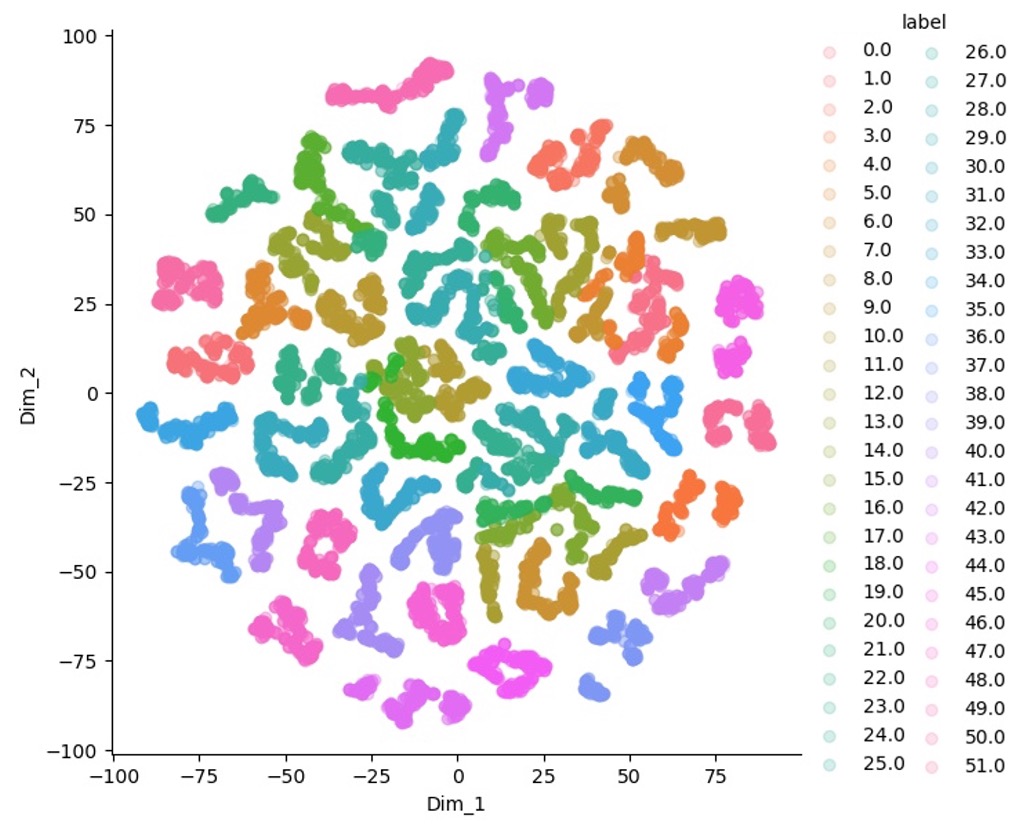}
            \caption{Trained end-to-end}
        \end{subfigure}
        \caption{Embed2Sym embedding spaces - Follow Suit 4 Standard.}
        \label{fig:e2s_fs_embed}
\end{figure}

In the Follow Suit domain, Embed2Sym* is able to cluster the playing card images in embedding space, but the normal end-to-end variant fails to separate the cards into 52 clusters. The TSNE plots for the embedding spaces for the first repeat of the 4-player standard variant, when the models are trained with 5300 examples, are shown in Figure \ref{fig:e2s_fs_embed}. Also, in both Embed2Sym variants, labelling the clusters is difficult, and to obtain results in a timely manner, we return the best cluster labelling the ASP Optimisation can find after 60 seconds. For reference, the cluster labelling in the MNIST Arithmetic domain completed in less than a second.

\begin{figure}[H]
    \centering
    \begin{subfigure}[t]{0.45\linewidth}
            \centering
            \includegraphics[width=1\linewidth]{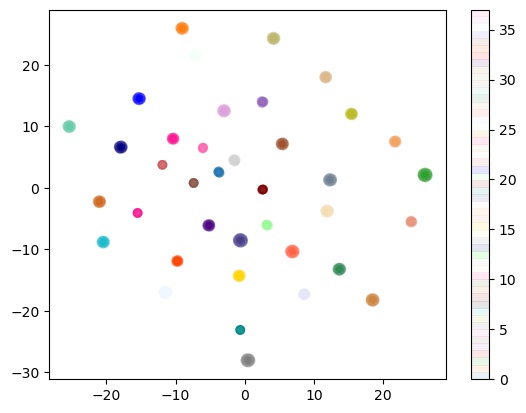}
            \caption{Before training}
        \end{subfigure}
        \begin{subfigure}[t]{0.45\linewidth}
            \centering
            \includegraphics[width=1\linewidth]{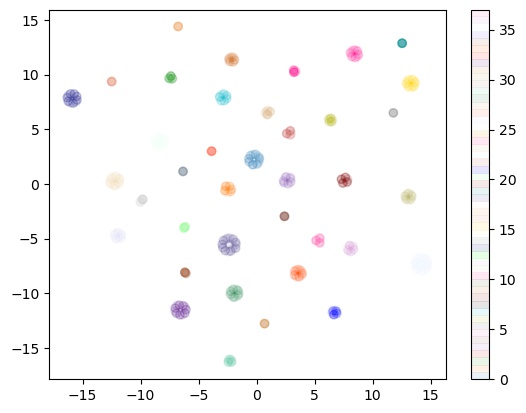}
            \caption{Trained with image labels}
        \end{subfigure}
        \begin{subfigure}[t]{0.45\linewidth}
            \includegraphics[width=\linewidth]{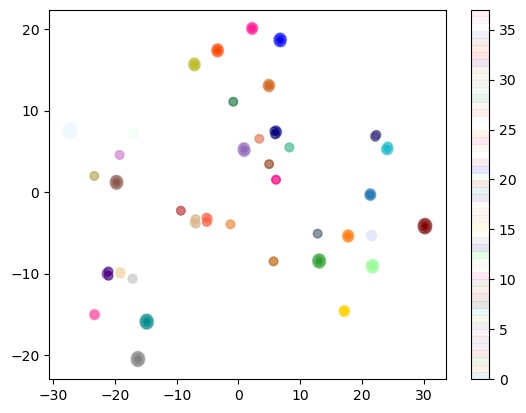}
            \caption{Trained end-to-end}
        \end{subfigure}
        \caption{Embed2Sym embedding spaces - Plant Hitting Sets.}
        \label{fig:e2s_phs_embed}
\end{figure}

In the Plants Hitting Sets domain, the ViT perception model is able to cluster the images without any fine-tuning, which explains why the results in both variants are fairly constant. Figure \ref{fig:e2s_phs_embed} shows the embedding space plots before training, and after training with both variants. In the before training case, we generate the TSNE plot with the 100-sample dataset that the cluster labelling stage observes, whereas after training, we use the images in the full dataset of 3800+100 samples. In the end-to-end case, the training data points in the embedding space are closer together. This makes it more difficult for k-means to identify general clusters in the data, to enable robust test-time performance. This explains why the performance of the Embed2Sym* variant is better than the end-to-end variant in Figure \ref{fig:plant_hs_reasoning}.

\textbf{SLASH}. We modified the optimiser to SGD with a learning rate of 0.01, as we obtained ``nan" values for the gradients. This matches the other baselines.

\textbf{BLIP+ABL}. To match our \method{} results, we train BLIP+ABL for 10 iterations on MNIST Arithmetic and Follow Suit, and 20 iterations on Plant Hitting Sets. On MNIST Arithmetic, we fine-tune BLIP for 10 epochs with a given set of abduced pseudo-labels each iteration. For Follow Suit and Plant Hitting Sets, we restricted the number of epochs to 1, but this did not prevent a timeout.

\textbf{NSIL}. In all experiments, we train the perception network for 1 epoch given the learned rules at each iteration. We also had to limit the number of answer sets to 5000 in each training sample, as in the Follow Suit domain there are too many corrective examples for ILASP. This was also required to train the network in NeurASP. In the MNIST Arithmetic and Follow Suit domains, we had to restrict NSIL to 10 iterations, in order to avoid a timeout. For Follow Suit, we had to further restrict the number of corrective examples by removing those examples with weights less than 5 (the lower the weight, the less influence the example has on the optimisation of the symbolic learner). Finally, we failed to obtain NSIL results in the Plant Hitting Sets domain, because the implementation does not support variable length training samples, and when attempting to run the initial bootstrap stage manually, the symbolic learner failed to return a hypothesis within a few hours. As NeurASP also timed out in this domain, we can assume NSIL would also reach a timeout, should the bootstrap task have succeeded.

\textbf{ViT+ReasoningNet}. We tried using an LSTM and a Recurrent Transformer as the reasoning network in the Follow Suit domains, but an MLP had superior performance. 

\textbf{BLIP+GPT4}. For this method, we use the same BLIP perception model as \method{}, and prompt GPT-4 to generate the symbolic component. For the Arithmetic and Follow Suit domains, we prompt GPT-4 to generate a python program, and for the Plant Hitting Sets domain GPT-4 generates an \ac{asp} program, as it is more natural to express the solution to this task in a declarative manner. We don't include the target rules directly in the prompt, but we do explain the various relations that can be used, to simulate passing the search space as occurs in the symbolic learner of \method{}. We also give substantial clues in the Plant Hitting Sets domain as to the correct rules, and provide context on the correct syntax of \ac{asp}. In all experiments we use the ``gpt-4" model from the OpenAI API, perform 5 repeats with 5 random seeds, and set the max tokens parameter to 5000. We access the API between 20th November - 30th December 2023. The prompts are shown in Listings \ref{lst:blip_gpt4_arith}, \ref{lst:blip_gpt4_fs}, and \ref{lst:blip_gpt4_phs} for the MNIST Arithmetic, Follow Suit, and Plant Hitting Sets domains respectively.

\begin{lstlisting}[caption={BLIP+GPT4 prompt - MNIST Arithmetic.}, label={lst:blip_gpt4_arith}]
The task is to generate a python program called 
"compute_answer" that returns a target answer, given a 
pair of digits as input. The program should generalise 
across a set of training examples and output the correct 
answer for all of the given input digit pairs. The digits 
are represented using a list, where the first and second 
elements denote the values of the first and second digit 
respectively. Note that the digits could be noisy as they 
are output from neural network predictions. The training 
examples are:

Example 1:
Input: [4, 8]
Output: 12

Example 2:
Input: [4, 6]
Output: 10

<<Insert more examples here>>

As background knowledge, consider using code that 
computes; (1)  whether a given digit is even or not, (2) 
the arithmetic sum "9 + the given digit", (3) arithmetic 
addition, and (4) equality. Note that the python function 
must generate exactly one output for a given pair of 
digits. The range of possible values for the answer is 
0-18, and for the input digits is 0-9, both inclusive. 
Generate the "compute_answer" python program that covers 
as many of the training examples as possible, whilst 
minimising the length of the program. Return your 
response as #### <python program> ####.
\end{lstlisting}

\begin{lstlisting}[caption={BLIP+GPT4 prompt - 4-player Follow Suit.}, label={lst:blip_gpt4_fs}]
The task is to generate a python program called 
"compute_answer" that returns a target answer for a card 
game, given a list of 4 playing cards as input. The 
cards in the list correspond to a card played by each of
4 players, and the answer denotes the winning player in 
the card game. Each player's card is denoted by the rank 
and then the suit, for example "kh" represents the card 
king of hearts, and "as" represents the card ace of 
spades. Note that for "10d", the rank is 10 and the suit 
is diamonds. The program should generalise across a set 
of training examples and output the correct answer for 
all of the given examples. Note that the cards could be 
noisy as they are output from neural network predictions. 
The training examples are:

Example 1:
Input: ['8s', '9d', '10s', '7d']
Output: 3

Example 2:
Input: ['8c', 'js', '10h', '6c']
Output: 1

<<Insert more examples here>>

As background knowledge, consider using code that 
computes; (1) whether the suit of one player's card is 
equal to a suit of another player's card, and (2), 
whether the rank of one player's card is higher than the
rank of another player's card. Note that the python 
function must generate exactly one output for a given 
list of cards. Consider the following python function 
that computes the value of a playing card rank, given as 
input a playing card. This can be used in the program:

def compute_rank_value(card):
    rank = card[:-1]
    if rank == 'j':
        rank_value = 11
    elif rank == 'q':
        rank_value = 12
    elif rank == 'k':
        rank_value = 13
    elif rank == 'a':
        rank_value = 14
    else:
        rank_value = int(rank)
    return rank_value

The range of possible values for the answer is 1-4, and 
assume the card inputs can be one of 52 playing cards 
from a standard deck. Generate the "compute_answer" 
python program that covers as many of the training 
examples as possible, whilst minimising the length of 
the program. Return your response as 
#### <python program> ####.
\end{lstlisting}

\begin{lstlisting}[caption={BLIP+GPT4 prompt - Plant Hitting Sets.}, label={lst:blip_gpt4_phs}]
The task is to generate an Answer Set Program (ASP) 
program that decides whether a given collection of sets 
contains a hitting set. A hitting set is a set that 
intersects with each set in the collection. Each set 
contains integer values, which are given as input. In 
this task, the hitting set is restricted to be of size 
2 or less. In each example, elements in each set are 
denoted with the predicate "ss_el", where the arguments 
correspond to the set identifier and the element itself. 
For example, "ss_el(1,5)." indicates that set 1 contains 
the element 5. The program should generalise across a 
set of training examples and return satisfiable or 
unsatisfiable (depending on whether the collection of 
elements contains a hitting set) for all of the given 
examples. Note the "ss_el" predicates could be noisy as 
the elements are generated from neural network 
predictions. The training examples are:

Example 1:
Input:
ss_el(1,24). ss_el(1,31). ss_el(1,33). ss_el(1,14). 
    ss_el(1,3). ss_el(2,26). ss_el(2,6). ss_el(2,28). 
    ss_el(2,37). ss_el(2,31). ss_el(3,29). ss_el(3,18). 
    ss_el(3,9). ss_el(4,10). ss_el(4,5). ss_el(4,24). 
    ss_el(4,26).
Has hitting set? No.

Example 2:
Input:
ss_el(1,36). ss_el(1,19). ss_el(1,31). ss_el(1,23). 
    ss_el(2,19). ss_el(2,35). ss_el(2,7). ss_el(2,27). 
    ss_el(2,32).
Has hitting set? Yes.

<<Insert more examples here>>

The following ASP background knowledge should be used in 
the program:

hs_index(1..2).
elt(1..38).
healthy(4). healthy(5). healthy(7). healthy(11). 
healthy(15). healthy(18). healthy(20). healthy(23). 
healthy(24). healthy(25). healthy(28). healthy(38).

The "hs_index" predicate denotes the index in the hitting 
set (1 or 2). "elt" denotes the range of possible integer 
elements. The "healthy" predicate indicates which integers 
have the healthy attribute and do not appear in the 
hitting set. When generating the ASP program, consider 
using the following predicates in the head of a rule: a 
"hs" predicate of arity 2 with the form "hs(X,Y)", which 
associates an element Y in the hitting set with it's 
index in the hitting set X, and a "hit" predicate of 
arity 1 with the form "hit(X)", to denote if a set X 
in the collection has been hit. In the body of the 
rule(s), consider using: (1) the "hs" predicate, (2) 
the != relation to denote that two elements are not 
equal, (3) the "ss_el" predicate, (4) the "hit" 
predicate, and (5) a "healthy" predicate to constrain 
the hitting set to only include non-healthy elements. 
You may use all necessary ASP constructs such as 
variables, negation, constraints and choice rules. In ASP,
negation of an atom p is expresses as "not p.". Negation 
of a predicate q(x) is expressed as "not q(x).". For 
choice rules, use the form "a { p(1); p(2); p(3)  } b."  
which is a disjunction and denotes "choose a minimum of 
a, and a maximum of b of elements from the set 
{ p(1), p(2), p(3) }". Don't use any weak constraints, 
optimisation statements, square brackets, or # statements. 
Include a constraint to ensure that the same elements 
occur at the same index in the hitting set. Also ensure 
that each set in the given collection is hit. Make sure 
all variables introduced in the head of a rule exist in 
the body, and that all variables are grounded by the 
background knowledge or the set elements. Try to generate 
an ASP program that covers as many of the training 
examples as possible, whilst minimising the length of 
the program. Return your response as 
#### <ASP program> ####.
\end{lstlisting}

It is also interesting to analyse the \ac{asp} programs GPT-4 generates. Listing \ref{lst:gpt4_phs_ex} shows an example where the correct rules are learned. This can be compared to the program learned by ILASP, presented in Figure \ref{fig:phs_rules}. GPT-4 generates the constraint that the hitting set must only contain healthy elements in the choice rule, which is a more compact representation than what ILASP learns (due to the representation given in the mode declarations). Although, GPT-4 repeats the healthy constraint in the third and fourth rules, which is redundant. GPT-4 also generates a constraint which states that the hitting set must not contain duplicate elements at the different indices, whereas ILASP learns the constraint that ensures different elements are not present at the same index. The third and fourth rules generated by GPT-4 are similar to what ILASP learns, as there is a definition of what it means for a set to be hit, and a constraint that ensures each set in the collection is hit. The other programs generated by GPT-4 are given in our GitHub repository.

\begin{lstlisting}[caption={Example GPT-4 output for Plant Hitting Sets.}, label={lst:gpt4_phs_ex}]
% Choice rule for hitting set elements
1 { hs(I,E) : hs_index(I), elt(E), not healthy(E) } 2.

% Constraint for same elements to occur at same index
:- hs(I,E1), hs(J,E2), I != J, E1 = E2.

% Each set should be hit by at least one element from the 
    hitting set
:- ss_el(S,E), not hit(S), elt(E), not healthy(E).

% A set is hit if it contains an element also in the 
    hitting set
hit(S) :- hs(I,E), ss_el(S,E), elt(E), not healthy(E).
\end{lstlisting}

\begin{table*}[t]
\centering
\resizebox{\textwidth}{!}{%
\begin{tabular}{@{}lrrrrrrrr@{}}
\cmidrule(l){2-8}
 & \multicolumn{2}{c}{\textbf{MNIST Arithmetic}} & \multicolumn{2}{c}{\textbf{Follow Suit 4}} & \multicolumn{2}{c}{\textbf{Follow Suit 10}} & \multicolumn{1}{c}{\textbf{Plant HS}} \\
 & \multicolumn{2}{c}{1100 labels} & \multicolumn{2}{c}{5300 labels} & \multicolumn{2}{c}{5300 labels} & \multicolumn{1}{c}{3900 labels} \\ \cmidrule(l){2-8}
                      & \multicolumn{1}{c}{\textbf{Sum}} & \multicolumn{1}{c}{\textbf{E9P}} & \multicolumn{1}{c}{\textbf{Standard}} & \multicolumn{1}{c}{\textbf{ACA}} & \multicolumn{1}{c}{\textbf{Standard}} & \multicolumn{1}{c}{\textbf{ACA}} & \multicolumn{1}{c}{}                  \\ \midrule
    
Embed2Sym* & 99.6 (1.5) & 111.7 (1.4) & 2322.6 (11.4) & 3132.3 (27.6) & 2343.1 (5.8) & 3302.6 (24.5) & 2030.9 (47.7) \\
Embed2Sym & 101.7 (0.3) & 101.2 (0.3) & 1136.2 (4.3) & 1291.3 (7.6) & 2859.5 (1.3) & 3382.2 (51.8) & 6626.2 (963.5) \\ 
ViT+ReasoningNet & 113.9 (0.3) & 118.2 (0.4) & 2534.8 (9.6) & 2693.3 (15.3) & 3160.1 (32.3) & 2568.4 (2.7) & 765.7 (11.3) \\ 
FF-NSL & 126.5 (0.5) & 126.9 (0.5) & 2382.4 (28.3) & 2645.4 (48.8) & 2399.0 (25.9) & 2660.9 (50.9) & 1247.0 (120.2) \\ 
NSIL & 545.2 (52.3) & 395.5 (25.8) & - & - & - & - & - \\ 
NeurASP & 689.4 (6.1) & 730.1 (8.2) & - & - & - & - & - \\ 
SLASH & 1112.6 (6.6) & 963.5 (27.4) & - & - & - & - & - \\ 
Ours (rules given) & 1557.6 (7.5) & 1557.6 (7.5) & 15597.4 (302.1) & 15147.6 (242.4) & 15597.4 (302.1) & 15147.6 (242.4) & 9653.1 (9.7) \\ 
BLIP+GPT4 & 1559.4 (7.7) & 1562.7 (7.4) & 15606.3 (301.8) & 15161.7 (242.4) & - & - & - \\ 
Ours & 1581.0 (8.0) & 1579.9 (7.5) & 15603.1 (301.8) & 15153.4 (242.2) & 15621.9 (302.8) & 15172.2 (241.3) & 9998.6 (85.7) \\ 
BLIP Single Image & 3495.1 (8.1) & 3530.7 (7.4) & 15747.1 (224.9) & 16033.7 (94.7) & 16346.4 (109.8) & 16378.9 (216.5) & 11478.4 (27.7) \\ 
BLIP+ABL & 17521.9 (290.8) & 17760.3 (326.1) & - & - & - & - & - \\ 

    \end{tabular}
    }
    \caption{Average learning time results (s)  when models are trained with the largest number of labels. Sorted by increasing learning time on the MNIST Arithmetic Sum task. Results show average time and standard error across 5 repeats.}
    \label{tab:learning_time}
    \end{table*}

\begin{figure}[H]
    \centering
    \begin{subfigure}[t]{0.3\linewidth}
            \centering
            \vspace{-5em}
            \includegraphics[width=1\linewidth]{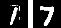}
            \caption{MNIST Arithmetic}
        \end{subfigure}
        \hfill
        \begin{subfigure}[t]{0.3\linewidth}
            \centering
            \includegraphics[width=1\linewidth]{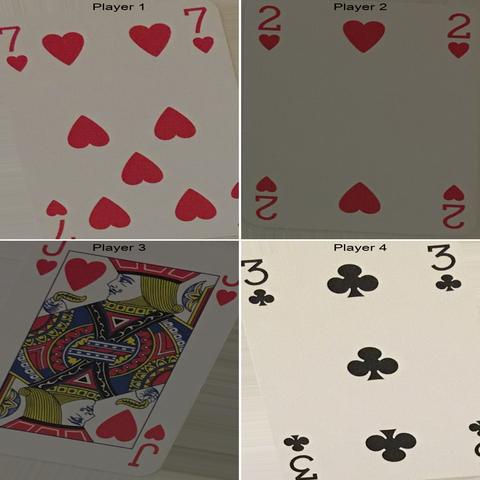}
            \caption{Follow Suit}
        \end{subfigure}
        \hfill
        \begin{subfigure}[t]{0.3\linewidth}
            \centering
            \includegraphics[width=1\linewidth]{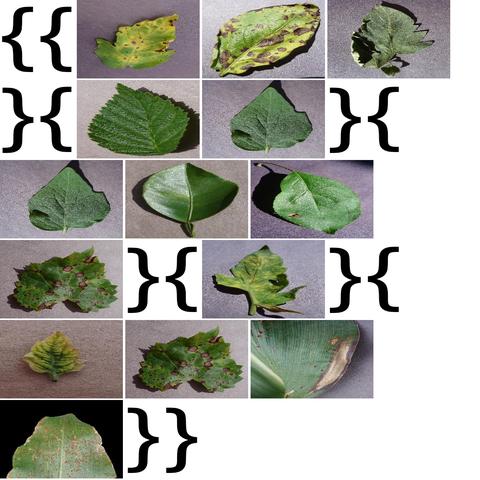}
            \caption{Plant Hitting Sets}
        \end{subfigure}
        \caption{BLIP Single Image examples.}
        \label{fig:blip_single_im}
\end{figure}

\textbf{BLIP Single Image}. For the Follow Suit domain, we include player annotations for each image, and for the Plant Hitting Sets domain, we encode the set assignment information as brackets within the image. All images are square as required by the BLIP image processors. Figure \ref{fig:blip_single_im} shows example data points for each domain.

\subsection{Learning Time}
The learning time results are shown in Table \ref{tab:learning_time}.

\subsection{Dataset Examples}
\begin{figure}[H]
    \centering
    \begin{subfigure}[t]{0.32\linewidth}
            \centering
            \includegraphics[width=1\linewidth]{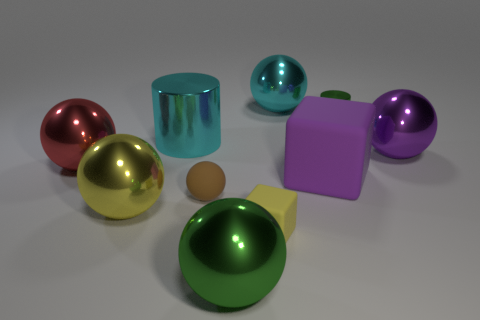}
            \caption{$y=1$}
        \end{subfigure}
        \hfill
        \begin{subfigure}[t]{0.32\linewidth}
            \centering
            \includegraphics[width=1\linewidth]{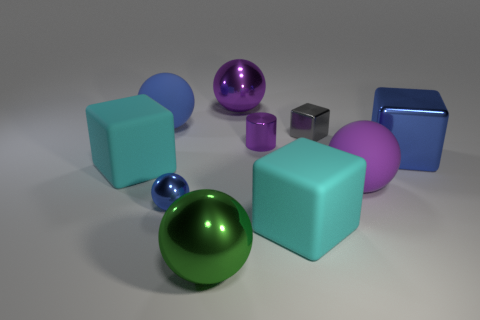}
            \caption{$y=2$}
        \end{subfigure}
        \hfill
        \begin{subfigure}[t]{0.32\linewidth}
            \centering
            \includegraphics[width=1\linewidth]{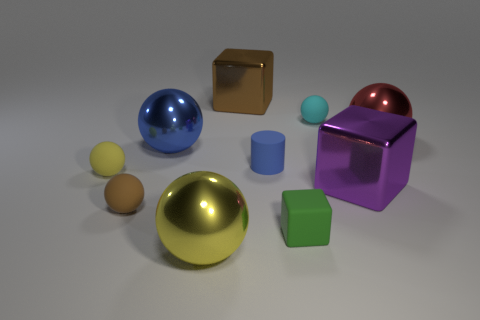}
            \caption{$y=3$}
        \end{subfigure}
        \caption{CLEVR-Hans3 example data points. The meta-data $M$ is empty. Class 1 images contain a large cube and a large cylinder. Class 2 images contain a small metal cube, and a small sphere. Class 3 images contain a large blue sphere, and a small yellow sphere.}
        \label{fig:clevr_hans_example}
\end{figure}

\begin{figure}[H]
    \centering
    \begin{subfigure}[t]{0.35\linewidth}
            \centering
            \includegraphics[width=1\linewidth]{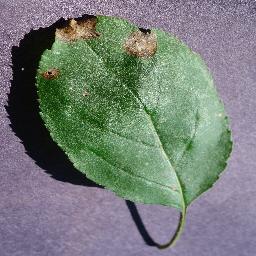}
            \caption{Apple Black Rot, Set 1}
        \end{subfigure}
        \hspace{2em}
        \begin{subfigure}[t]{0.35\linewidth}
            \centering
            \includegraphics[width=1\linewidth]{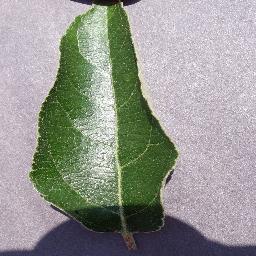}
            \caption{Apple Healthy, Set 2}
        \end{subfigure}
        
        \begin{subfigure}[t]{0.35\linewidth}
            \centering
            \includegraphics[width=1\linewidth]{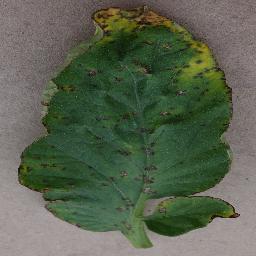}
            \caption{Tomato Bacterial Spot, Set 2}
        \end{subfigure}
        \hspace{2em}
        \begin{subfigure}[t]{0.35\linewidth}
            \centering
            \includegraphics[width=1\linewidth]{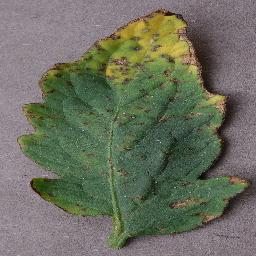}
            \caption{Tomato Bacterial Spot, Set 3}
        \end{subfigure}
        \caption{Plant Hitting Sets example data point. The meta-data $M$ indicates which set each element belongs to. The label is $y=1$ as there is a hitting set of size $\leq$ 2 with the elements Apple Black Rot, and Tomato Bacterial Spot.}
        \label{fig:phs_example}
\end{figure}

In addition to the Follow Suit data point presented in Figure \ref{fig:arch} in the main paper, Figures \ref{fig:clevr_hans_example}, and \ref{fig:phs_example} show example data points in the CLEVR-Hans3, and Plant Hitting Sets tasks respectively.


\subsection{Zero-shot Evaluation}
We consider whether recent vision-language foundation models such as BLIP2 \cite{li2023blip}, and GPT-4 vision (GPT-4V) \cite{gpt4v} can be used out-of-the-box within the \method{} architecture. These models don't currently support fine-tuning with VQA, but they do support VQA inference, and they are arguably more advanced than the original BLIP approach we use in our main results (BLIP1). To investigate this, we generate image predictions in a zero-shot fashion via natural language questions or prompts, and then use these image predictions to perform rule learning in our downstream tasks. When performing rule learning, we present the mean task accuracy achieved over 5 repeats.

With both BLIP approaches, we consider asking the models to generate the image predictions directly, and on the Follow Suit and Plant Hitting Sets images, via a conditional questioning approach (denoted CQA). This is useful when the image features can easily be decomposed into various questions. For example, in the playing cards domain, one can initially ask the question ``which color are the symbols?", before asking ``which symbol is on the card". Candidate answers can then be restricted accordingly, e.g., to force the model to return ``hearts" or ``diamonds" for the symbol question, if the color is predicted as ``red". We don't use this technique for the MNIST images as these are challenging to decompose.

We also evaluate the ``generate" and ``rank" inference modes in BLIP1, where the ``generate" approach is the same as the one presented in the main body of this paper, with the string distance measure included. The rank approach requires giving possible candidate answers to BLIP, which then returns the answer that minimises the language model loss between each candidate and the generated output. We didn't use this for \method{} in the main results, because after fine-tuning, we obtained better results using the generate mode. For BLIP2, we evaluated the Flan-t5 XXL instruct, feature extractor, and Image-Text-Matching (ITM) modes, and selected the ITM mode as this resulted in the best performance. For GPT-4V, we use the OpenAI API with the model ``gpt-4-vision-preview", and set the max tokens parameter to 300. All responses were obtained when accessing the API between 19th-30th December 2023. Finally, in all tasks, we used 100 downstream examples.

\begin{table}[t]
\centering
\resizebox{1\linewidth}{!}{%
\begin{tabular}{@{}lrrrr@{}}
\cmidrule(l){2-5}
 & \multicolumn{1}{c}{\textbf{MNIST}} & \multicolumn{2}{c}{\textbf{Playing Cards}} & \multicolumn{1}{c}{\textbf{Plant Disease}} \\ \cmidrule(l){2-5}
\multicolumn{1}{l}{} &  & \multicolumn{1}{c}{\textbf{Standard}} & \multicolumn{1}{c}{\textbf{ACA}} & \\ \midrule
    
\multicolumn{1}{l}{BLIP1 Generate} & 0.27 & 0.074 & 0.0096 & 0.0008 \\
\multicolumn{1}{l}{BLIP1 Rank} & 0.2262 & 0.0971 & 0.0567 & 0.035 \\
\multicolumn{1}{l}{BLIP1 Rank CQA} & - & 0.6962 & 0.225 & 0.0326 \\ \hline
\multicolumn{1}{l}{BLIP2 ITM} & \textbf{0.6521} & \textbf{0.7846} & \textbf{0.3096} & 0.1214 \\
\multicolumn{1}{l}{BLIP2 ITM CQA} & - & 0.6385 & 0.301 & \textbf{0.1624} \\ \hline
\multicolumn{1}{l}{GPT-4 Vision$^*$} & 0.5178 & \textbf{0.9048} & \textbf{0.8625} & 0.04 \\

\end{tabular}
}
\caption{Foundation model zero-shot image accuracy.}
\label{tab:zero_image_acc}
\end{table}

\begin{table}[t]
\centering
\resizebox{1\linewidth}{!}{%
\begin{tabular}{@{}lrrrrrrrr@{}}
\cmidrule(l){2-8}
& \multicolumn{2}{c}{\textbf{Arithmetic}} & \multicolumn{2}{c}{\textbf{Follow Suit 4}} & \multicolumn{2}{c}{\textbf{Follow Suit 10}} & \textbf{Plant HS} \\ \cmidrule(l){2-8} 
\multicolumn{1}{l}{}            & \multicolumn{1}{l}{\textbf{Sum}}  & \textbf{E9P} & \textbf{Standard}           & \textbf{ACA}          & \textbf{Standard} & \multicolumn{1}{c}{\textbf{ACA}} & \\ \midrule

\multicolumn{1}{l}{BLIP1 Generate} & 0.089 & 0.1484 & 0.4198 & 0.3038 & 0.174 & 0.1226 & 0.7688 \\
\multicolumn{1}{l}{BLIP1 Rank} & 0.0742 & 0.1018 & 0.301 & 0.2806 & 0.1784 & 0.1716 & 0.7624 \\
\multicolumn{1}{l}{BLIP1 Rank CQA} & - & - & 0.4278 & 0.2914 & 0.271 & 0.0808 & 0.7176 \\ \hline
\multicolumn{1}{l}{BLIP2 ITM} & \textbf{0.4276} & \textbf{0.5458} & \textbf{0.5074} & \textbf{0.3734} & \textbf{0.3168} & \textbf{0.1854} & 0.7252 \\
\multicolumn{1}{l}{BLIP2 ITM CQA} & - & - & 0.4706 & 0.2858 & 0.2688 & 0.115 & \textbf{0.8102} \\ \hline
\multicolumn{1}{l}{GPT-4 Vision$^*$} & 0.2602 & 0.2982 & 0.4502 & \textbf{0.4886} & 0.2026 & \textbf{0.235} & - \\

    \end{tabular}
    }
    \caption{Foundation model task accuracy with zero-shot image predictions.}
    \label{tab:zero_task_acc}
    \end{table}
    
The image and downstream task accuracy results are shown in Tables \ref{tab:zero_image_acc} and \ref{tab:zero_task_acc} respectively. As you can see, more recent models such as BLIP2 and GPT-4V outperform BLIP1, but still fall significantly below the levels of accuracy achieved in the main body of the paper, where a small amount of fine-tuning was used with labelled data. Also, the correct rules are only learned on the Arithmetic tasks when BLIP2 and GPT-4V are used zero-shot, but again, the task accuracy is lower than that presented in Figure \ref{fig:learning} in the main paper. 

The * indicates that GPT-4V was unable to return a classification for all of the images in the datasets.\footnote{We often observe error responses from the OpenAI API, including file extension errors, and errors regarding violation of the profanity filter.} The error rates are 38.83\% (MNIST), 8.08\% (Playing Cards Standard), 4.23\% (Playing Cards ACA), and 92.16\% (Plant Disease). Note that on the Plant Disease dataset, most of the errors were GPT-4V refusing to make a classification, citing a lack of information, rather than an error from the API. The accuracy results presented in Table \ref{tab:zero_image_acc} are measured over the full test set, and therefore include erroneous data points. When rule learning is performed, examples without a complete set of image predictions are removed from the symbolic learning task. On the Plant Hitting Sets task with GPT-4V, there are therefore no examples and no rules can be learned. If a test example has missing image predictions, it is assumed incorrect.

The best results are on the Follow Suit tasks, although even with an image accuracy of 0.9048 on the standard deck, it is insufficient to learn the correct rules. With more downstream task examples, it may be possible to do so and improve the task accuracy. These results demonstrate that the fine-tuning approach of \method{} is currently required to achieve strong performance. In future, should BLIP2 or GPT-4V support VQA fine-tuning, it may be possible to improve upon the \method{} results presented in the main body of this paper.
\subsection{Full Results}
Tables \ref{tab:full_arithmetic} - \ref{tab:full_plant_hs} show the full \method{} results for all data points in all domains.

\begin{table*}[t]
    \centering
    \resizebox{0.8\textwidth}{!}{%
    \begin{tabular}{@{}cccrrrr@{}}
    \cmidrule(l){4-7}
    \multicolumn{1}{l}{}                  & \multicolumn{1}{l}{}                       & \multicolumn{1}{l}{}           & \multicolumn{2}{c}{\textbf{Sum}}                                       & \multicolumn{2}{c}{\textbf{E9P}}                                       \\ \midrule
    \multicolumn{1}{c}{\textbf{Num. Image Labels}} & \multicolumn{1}{c}{\textbf{Num. Task Labels}} & \multicolumn{1}{c}{\textbf{Image Acc.}} & \multicolumn{1}{c}{\textbf{Rule Acc.}} & \multicolumn{1}{c}{\textbf{Task Acc.}} & \multicolumn{1}{c}{\textbf{Rule Acc.}} & \multicolumn{1}{c}{\textbf{Task Acc.}} \\ \midrule
    \multirow{4}{*}{0} & 10 & \multirow{4}{*}{0.2700} & 0.19 (0.11) & 0.07 (0.01) & 0.30 (0.09) & 0.10 (0.02)  \\
 & 20 & & 0.25 (0.09) & 0.07 (0.01) & 0.35 (0.08) & 0.12 (0.02)  \\
 & 50 & & 0.19 (0.10) & 0.09 (0.00) & 0.40 (0.09) & 0.14 (0.01)  \\
 & 100 & & 0.16 (0.06) & 0.09 (0.00) & 0.52 (0.01) & 0.15 (0.00) \\ \midrule
\multirow{4}{*}{10} & 10 & \multirow{4}{*}{0.7217 (0.0130)} & 0.70 (0.17) & 0.39 (0.09) & 0.86 (0.09) & 0.56 (0.05)  \\
 & 20 & & 0.92 (0.08) & 0.48 (0.04) & 1.00 (0.00) & 0.65 (0.01)  \\
 & 50 & & 1.00 (0.00) & 0.53 (0.02) & 1.00 (0.00) & 0.65 (0.01)  \\
 & 100 & & 1.00 (0.00) & 0.53 (0.02) & 1.00 (0.00) & 0.65 (0.01) \\ \midrule
\multirow{4}{*}{20} & 10 & \multirow{4}{*}{0.7891 (0.0139)} & 0.87 (0.09) & 0.55 (0.06) & 0.91 (0.05) & 0.67 (0.04)  \\
 & 20 & & 0.96 (0.04) & 0.60 (0.03) & 1.00 (0.00) & 0.74 (0.01)  \\
 & 50 & & 1.00 (0.00) & 0.63 (0.02) & 1.00 (0.00) & 0.74 (0.01)  \\
 & 100 & & 1.00 (0.00) & 0.63 (0.02) & 1.00 (0.00) & 0.74 (0.01) \\ \midrule
\multirow{4}{*}{50} & 10 & \multirow{4}{*}{0.8849 (0.0036)} & 0.92 (0.05) & 0.71 (0.04) & 0.95 (0.05) & 0.81 (0.04)  \\
 & 20 & & 0.91 (0.06) & 0.71 (0.04) & 1.00 (0.00) & 0.85 (0.01)  \\
 & 50 & & 1.00 (0.00) & 0.78 (0.01) & 1.00 (0.00) & 0.85 (0.01)  \\
 & 100 & & 1.00 (0.00) & 0.78 (0.01) & 1.00 (0.00) & 0.85 (0.01) \\ \midrule
\multirow{4}{*}{100} & 10 & \multirow{4}{*}{0.9216 (0.0024)} & 1.00 (0.00) & 0.85 (0.01) & 1.00 (0.00) & 0.90 (0.00)  \\
 & 20 & & 1.00 (0.00) & 0.85 (0.01) & 1.00 (0.00) & 0.90 (0.00)  \\
 & 50 & & 1.00 (0.00) & 0.85 (0.01) & 1.00 (0.00) & 0.90 (0.00)  \\
 & 100 & & 1.00 (0.00) & 0.85 (0.01) & 1.00 (0.00) & 0.90 (0.00) \\ \midrule
\multirow{4}{*}{200} & 10 & \multirow{4}{*}{0.9485 (0.0030)} & 1.00 (0.00) & 0.90 (0.00) & 1.00 (0.00) & 0.93 (0.00)  \\
 & 20 & & 1.00 (0.00) & 0.90 (0.00) & 1.00 (0.00) & 0.93 (0.00)  \\
 & 50 & & 1.00 (0.00) & 0.90 (0.00) & 1.00 (0.00) & 0.93 (0.00)  \\
 & 100 & & 1.00 (0.00) & 0.90 (0.00) & 1.00 (0.00) & 0.93 (0.00) \\ \midrule
\multirow{4}{*}{500} & 10 & \multirow{4}{*}{0.9685 (0.0010)} & 1.00 (0.00) & 0.94 (0.00) & 1.00 (0.00) & 0.96 (0.00)  \\
 & 20 & & 1.00 (0.00) & 0.94 (0.00) & 1.00 (0.00) & 0.96 (0.00)  \\
 & 50 & & 1.00 (0.00) & 0.94 (0.00) & 1.00 (0.00) & 0.96 (0.00)  \\
 & 100 & & 1.00 (0.00) & 0.94 (0.00) & 1.00 (0.00) & 0.96 (0.00) \\ \midrule
\multirow{4}{*}{1000} & 10 & \multirow{4}{*}{0.9753 (0.0013)} & 1.00 (0.00) & 0.95 (0.00) & 1.00 (0.00) & 0.96 (0.00)  \\
 & 20 & & 1.00 (0.00) & 0.95 (0.00) & 1.00 (0.00) & 0.96 (0.00)  \\
 & 50 & & 1.00 (0.00) & 0.95 (0.00) & 1.00 (0.00) & 0.96 (0.00)  \\
 & 100 & & 1.00 (0.00) & 0.95 (0.00) & 1.00 (0.00) & 0.96 (0.00) \\ \bottomrule

    \end{tabular}
    }
    \caption{Full MNIST Arithmetic results.}
    \label{tab:full_arithmetic}
    \end{table*}

\begin{table*}[t]
    \centering
    \resizebox{0.8\textwidth}{!}{%
    \begin{tabular}{@{}cccrrrr@{}}
    \cmidrule(l){4-7}
    \multicolumn{1}{l}{}                  & \multicolumn{1}{l}{}                       & \multicolumn{1}{l}{}           & \multicolumn{2}{c}{\textbf{4 Players}}                                       & \multicolumn{2}{c}{\textbf{10 Players}}                                       \\ \midrule
    \multicolumn{1}{c}{\textbf{Num. Image Labels}} & \multicolumn{1}{c}{\textbf{Num. Task Labels}} & \multicolumn{1}{c}{\textbf{Image Acc.}} & \multicolumn{1}{c}{\textbf{Rule Acc.}} & \multicolumn{1}{c}{\textbf{Task Acc.}} & \multicolumn{1}{c}{\textbf{Rule Acc.}} & \multicolumn{1}{c}{\textbf{Task Acc.}} \\ \midrule
    \multirow{4}{*}{0} & 10 & \multirow{4}{*}{0.0740} & 0.55 (0.01) & 0.45 (0.01) & 0.15 (0.07) & 0.15 (0.05)  \\
 & 20 & & 0.47 (0.07) & 0.43 (0.01) & 0.22 (0.09) & 0.10 (0.03)  \\
 & 50 & & 0.37 (0.07) & 0.41 (0.01) & 0.19 (0.06) & 0.14 (0.04)  \\
 & 100 & & 0.78 (0.14) & 0.42 (0.01) & 0.19 (0.06) & 0.17 (0.03) \\ \midrule
\multirow{4}{*}{52} & 10 & \multirow{4}{*}{0.9833 (0.0106)} & 1.00 (0.00) & 1.00 (0.00) & 1.00 (0.00) & 1.00 (0.00)  \\
 & 20 & & 1.00 (0.00) & 1.00 (0.00) & 1.00 (0.00) & 1.00 (0.00)  \\
 & 50 & & 1.00 (0.00) & 1.00 (0.00) & 1.00 (0.00) & 1.00 (0.00)  \\
 & 100 & & 1.00 (0.00) & 1.00 (0.00) & 1.00 (0.00) & 1.00 (0.00) \\ \midrule
\multirow{4}{*}{104} & 10 & \multirow{4}{*}{0.9988 (0.0009)} & 1.00 (0.00) & 1.00 (0.00) & 1.00 (0.00) & 1.00 (0.00)  \\
 & 20 & & 1.00 (0.00) & 1.00 (0.00) & 1.00 (0.00) & 1.00 (0.00)  \\
 & 50 & & 1.00 (0.00) & 1.00 (0.00) & 1.00 (0.00) & 1.00 (0.00)  \\
 & 100 & & 1.00 (0.00) & 1.00 (0.00) & 1.00 (0.00) & 1.00 (0.00) \\ \midrule
\multirow{4}{*}{260} & 10 & \multirow{4}{*}{0.9983 (0.0015)} & 1.00 (0.00) & 1.00 (0.00) & 1.00 (0.00) & 1.00 (0.00)  \\
 & 20 & & 1.00 (0.00) & 1.00 (0.00) & 1.00 (0.00) & 1.00 (0.00)  \\
 & 50 & & 1.00 (0.00) & 1.00 (0.00) & 1.00 (0.00) & 1.00 (0.00)  \\
 & 100 & & 1.00 (0.00) & 1.00 (0.00) & 1.00 (0.00) & 1.00 (0.00) \\ \midrule
\multirow{4}{*}{520} & 10 & \multirow{4}{*}{1.0000 (0.0000)} & 1.00 (0.00) & 1.00 (0.00) & 1.00 (0.00) & 1.00 (0.00)  \\
 & 20 & & 1.00 (0.00) & 1.00 (0.00) & 1.00 (0.00) & 1.00 (0.00)  \\
 & 50 & & 1.00 (0.00) & 1.00 (0.00) & 1.00 (0.00) & 1.00 (0.00)  \\
 & 100 & & 1.00 (0.00) & 1.00 (0.00) & 1.00 (0.00) & 1.00 (0.00) \\ \midrule
\multirow{4}{*}{1040} & 10 & \multirow{4}{*}{1.0000 (0.0000)} & 1.00 (0.00) & 1.00 (0.00) & 1.00 (0.00) & 1.00 (0.00)  \\
 & 20 & & 1.00 (0.00) & 1.00 (0.00) & 1.00 (0.00) & 1.00 (0.00)  \\
 & 50 & & 1.00 (0.00) & 1.00 (0.00) & 1.00 (0.00) & 1.00 (0.00)  \\
 & 100 & & 1.00 (0.00) & 1.00 (0.00) & 1.00 (0.00) & 1.00 (0.00) \\ \midrule
\multirow{4}{*}{2600} & 10 & \multirow{4}{*}{1.0000 (0.0000)} & 1.00 (0.00) & 1.00 (0.00) & 1.00 (0.00) & 1.00 (0.00)  \\
 & 20 & & 1.00 (0.00) & 1.00 (0.00) & 1.00 (0.00) & 1.00 (0.00)  \\
 & 50 & & 1.00 (0.00) & 1.00 (0.00) & 1.00 (0.00) & 1.00 (0.00)  \\
 & 100 & & 1.00 (0.00) & 1.00 (0.00) & 1.00 (0.00) & 1.00 (0.00) \\ \midrule
\multirow{4}{*}{5200} & 10 & \multirow{4}{*}{1.0000 (0.0000)} & 1.00 (0.00) & 1.00 (0.00) & 1.00 (0.00) & 1.00 (0.00)  \\
 & 20 & & 1.00 (0.00) & 1.00 (0.00) & 1.00 (0.00) & 1.00 (0.00)  \\
 & 50 & & 1.00 (0.00) & 1.00 (0.00) & 1.00 (0.00) & 1.00 (0.00)  \\
 & 100 & & 1.00 (0.00) & 1.00 (0.00) & 1.00 (0.00) & 1.00 (0.00) \\ \bottomrule

    \end{tabular}
    }
    \caption{Full Follow Suit results, standard deck.}
    \label{tab:full_fs_standard}
    \end{table*}
    
\begin{table*}[t]
\centering
\resizebox{0.8\textwidth}{!}{%
\begin{tabular}{@{}cccrrrr@{}}
\cmidrule(l){4-7}
\multicolumn{1}{l}{}                  & \multicolumn{1}{l}{}                       & \multicolumn{1}{l}{}           & \multicolumn{2}{c}{\textbf{4 Players}}                                       & \multicolumn{2}{c}{\textbf{10 Players}}                                       \\ \midrule
\multicolumn{1}{c}{\textbf{Num. Image Labels}} & \multicolumn{1}{c}{\textbf{Num. Task Labels}} & \multicolumn{1}{c}{\textbf{Image Acc.}} & \multicolumn{1}{c}{\textbf{Rule Acc.}} & \multicolumn{1}{c}{\textbf{Task Acc.}} & \multicolumn{1}{c}{\textbf{Rule Acc.}} & \multicolumn{1}{c}{\textbf{Task Acc.}} \\ \midrule
\multirow{4}{*}{0} & 10 & \multirow{4}{*}{0.0096} & 0.10 (0.05) & 0.24 (0.03) & 0.05 (0.05) & 0.08 (0.02)  \\
 & 20 & & 0.06 (0.04) & 0.29 (0.01) & 0.03 (0.02) & 0.13 (0.00)  \\
 & 50 & & 0.03 (0.03) & 0.27 (0.03) & 0.02 (0.01) & 0.12 (0.01)  \\
 & 100 & & 0.04 (0.03) & 0.30 (0.00) & 0.05 (0.02) & 0.12 (0.01) \\ \midrule
\multirow{4}{*}{52} & 10 & \multirow{4}{*}{0.9988 (0.0006)} & 1.00 (0.00) & 1.00 (0.00) & 1.00 (0.00) & 1.00 (0.00)  \\
 & 20 & & 1.00 (0.00) & 1.00 (0.00) & 1.00 (0.00) & 1.00 (0.00)  \\
 & 50 & & 1.00 (0.00) & 1.00 (0.00) & 1.00 (0.00) & 1.00 (0.00)  \\
 & 100 & & 1.00 (0.00) & 1.00 (0.00) & 1.00 (0.00) & 1.00 (0.00) \\ \midrule
\multirow{4}{*}{104} & 10 & \multirow{4}{*}{0.9994 (0.0004)} & 1.00 (0.00) & 1.00 (0.00) & 1.00 (0.00) & 1.00 (0.00)  \\
 & 20 & & 1.00 (0.00) & 1.00 (0.00) & 1.00 (0.00) & 1.00 (0.00)  \\
 & 50 & & 1.00 (0.00) & 1.00 (0.00) & 1.00 (0.00) & 1.00 (0.00)  \\
 & 100 & & 1.00 (0.00) & 1.00 (0.00) & 1.00 (0.00) & 1.00 (0.00) \\ \midrule
\multirow{4}{*}{260} & 10 & \multirow{4}{*}{0.9996 (0.0002)} & 1.00 (0.00) & 1.00 (0.00) & 1.00 (0.00) & 1.00 (0.00)  \\
 & 20 & & 1.00 (0.00) & 1.00 (0.00) & 1.00 (0.00) & 1.00 (0.00)  \\
 & 50 & & 1.00 (0.00) & 1.00 (0.00) & 1.00 (0.00) & 1.00 (0.00)  \\
 & 100 & & 1.00 (0.00) & 1.00 (0.00) & 1.00 (0.00) & 1.00 (0.00) \\ \midrule
\multirow{4}{*}{520} & 10 & \multirow{4}{*}{0.9996 (0.0002)} & 1.00 (0.00) & 1.00 (0.00) & 1.00 (0.00) & 1.00 (0.00)  \\
 & 20 & & 1.00 (0.00) & 1.00 (0.00) & 1.00 (0.00) & 1.00 (0.00)  \\
 & 50 & & 1.00 (0.00) & 1.00 (0.00) & 1.00 (0.00) & 1.00 (0.00)  \\
 & 100 & & 1.00 (0.00) & 1.00 (0.00) & 1.00 (0.00) & 1.00 (0.00) \\ \midrule
\multirow{4}{*}{1040} & 10 & \multirow{4}{*}{0.9996 (0.0002)} & 1.00 (0.00) & 1.00 (0.00) & 1.00 (0.00) & 1.00 (0.00)  \\
 & 20 & & 1.00 (0.00) & 1.00 (0.00) & 1.00 (0.00) & 1.00 (0.00)  \\
 & 50 & & 1.00 (0.00) & 1.00 (0.00) & 1.00 (0.00) & 1.00 (0.00)  \\
 & 100 & & 1.00 (0.00) & 1.00 (0.00) & 1.00 (0.00) & 1.00 (0.00) \\ \midrule
\multirow{4}{*}{2600} & 10 & \multirow{4}{*}{1.0000 (0.0000)} & 1.00 (0.00) & 1.00 (0.00) & 1.00 (0.00) & 1.00 (0.00)  \\
 & 20 & & 1.00 (0.00) & 1.00 (0.00) & 1.00 (0.00) & 1.00 (0.00)  \\
 & 50 & & 1.00 (0.00) & 1.00 (0.00) & 1.00 (0.00) & 1.00 (0.00)  \\
 & 100 & & 1.00 (0.00) & 1.00 (0.00) & 1.00 (0.00) & 1.00 (0.00) \\ \midrule
\multirow{4}{*}{5200} & 10 & \multirow{4}{*}{1.0000 (0.0000)} & 1.00 (0.00) & 1.00 (0.00) & 1.00 (0.00) & 1.00 (0.00)  \\
 & 20 & & 1.00 (0.00) & 1.00 (0.00) & 1.00 (0.00) & 1.00 (0.00)  \\
 & 50 & & 1.00 (0.00) & 1.00 (0.00) & 1.00 (0.00) & 1.00 (0.00)  \\
 & 100 & & 1.00 (0.00) & 1.00 (0.00) & 1.00 (0.00) & 1.00 (0.00) \\ \bottomrule

    \end{tabular}
    }
    \caption{Full Follow Suit results, ACA deck.}
    \label{tab:full_fs_aca}
    \end{table*}

\begin{table*}[t]
    \centering
    \resizebox{0.7\textwidth}{!}{%
    \begin{tabular}{@{}cccrr@{}}
    \midrule
    \multicolumn{1}{c}{\textbf{Num. Image Labels}} & \multicolumn{1}{c}{\textbf{Num. Task Labels}} & \multicolumn{1}{c}{\textbf{Image Acc.}} & \multicolumn{1}{c}{\textbf{Rule Acc.}} & \multicolumn{1}{c}{\textbf{Task Acc.}} \\ \midrule
    \multirow{4}{*}{0} & 10 & \multirow{4}{*}{0.0008 (0.0000)} & 0.77 (0.06) & 0.71 (0.02)  \\
 & 20 & & 0.75 (0.07) & 0.69 (0.03)  \\
 & 50 & & 0.74 (0.02) & 0.75 (0.02)  \\
 & 100 & & 0.81 (0.05) & 0.77 (0.00) \\ \midrule
\multirow{4}{*}{38} & 10 & \multirow{4}{*}{0.2934 (0.0104)} & 0.94 (0.05) & 0.82 (0.03)  \\
 & 20 & & 0.95 (0.04) & 0.83 (0.02)  \\
 & 50 & & 0.94 (0.04) & 0.84 (0.02)  \\
 & 100 & & 0.97 (0.02) & 0.86 (0.01) \\ \midrule
\multirow{4}{*}{76} & 10 & \multirow{4}{*}{0.6016 (0.0125)} & 0.88 (0.07) & 0.84 (0.04)  \\
 & 20 & & 0.98 (0.01) & 0.91 (0.01)  \\
 & 50 & & 0.98 (0.01) & 0.91 (0.01)  \\
 & 100 & & 0.98 (0.01) & 0.91 (0.01) \\ \midrule
\multirow{4}{*}{190} & 10 & \multirow{4}{*}{0.7586 (0.0025)} & 0.98 (0.01) & 0.92 (0.01)  \\
 & 20 & & 0.95 (0.04) & 0.90 (0.03)  \\
 & 50 & & 0.97 (0.01) & 0.92 (0.01)  \\
 & 100 & & 0.98 (0.01) & 0.93 (0.00) \\ \midrule
\multirow{4}{*}{380} & 10 & \multirow{4}{*}{0.8826 (0.0058)} & 0.93 (0.04) & 0.91 (0.04)  \\
 & 20 & & 0.99 (0.01) & 0.96 (0.01)  \\
 & 50 & & 0.99 (0.01) & 0.96 (0.01)  \\
 & 100 & & 1.00 (0.00) & 0.97 (0.00) \\ \midrule
\multirow{4}{*}{760} & 10 & \multirow{4}{*}{0.9456 (0.0030)} & 0.92 (0.04) & 0.91 (0.04)  \\
 & 20 & & 0.95 (0.04) & 0.94 (0.04)  \\
 & 50 & & 0.99 (0.01) & 0.98 (0.01)  \\
 & 100 & & 1.00 (0.00) & 0.98 (0.00) \\ \midrule
\multirow{4}{*}{1900} & 10 & \multirow{4}{*}{0.9784 (0.0011)} & 0.93 (0.04) & 0.93 (0.04)  \\
 & 20 & & 0.99 (0.01) & 0.98 (0.01)  \\
 & 50 & & 0.99 (0.01) & 0.99 (0.01)  \\
 & 100 & & 1.00 (0.00) & 0.99 (0.00) \\ \midrule
\multirow{4}{*}{3800} & 10 & \multirow{4}{*}{0.9871 (0.0008)} & 0.93 (0.04) & 0.93 (0.04)  \\
 & 20 & & 0.99 (0.01) & 0.99 (0.01)  \\
 & 50 & & 0.99 (0.01) & 0.99 (0.01)  \\
 & 100 & & 1.00 (0.00) & 1.00 (0.00) \\ \bottomrule

    \end{tabular}
    }
    \caption{Full Plant Hitting Sets results.}
    \label{tab:full_plant_hs}
    \end{table*}

\end{document}